\documentclass[runningheads]{llncs}

\usepackage{eccv}

\usepackage{eccvabbrv}

\usepackage{graphicx}
\usepackage{booktabs, multicol, multirow}
\usepackage{tcolorbox}
\usepackage{pifont, tabularx, caption}
\usepackage{makecell}
 
\usepackage[accsupp]{axessibility}  

\usepackage[pagebackref,breaklinks,colorlinks,citecolor=eccvblue]{hyperref}
\usepackage{hyperref}

\usepackage{orcidlink}

\begin{document}

\title{ToG-Bench: Task-Oriented Spatio-Temporal Grounding in Egocentric Videos}

\titlerunning{ToG-Bench}


%
\author{
  Qi'ao Xu\textsuperscript{1}, 
  Tianwen Qian\textsuperscript{1},
  Yuqian Fu\textsuperscript{2}, 
  Kailing Li\textsuperscript{1}, 
  Yang Jiao\textsuperscript{3},\\ 
  Jiacheng Zhang\textsuperscript{3}, 
  Xiaoling Wang\textsuperscript{1\dag}, 
  Liang He\textsuperscript{1}
}

\institute{
  East China Normal University, Shanghai, China \\
  \email{qaxu@stu.ecnu.edu.cn, \{twqian,xlwang\}@cs.ecnu.edu.cn}
  \and
  INSAIT, Sofia University ``St. Kliment Ohridski'', Sofia, Bulgaria
  \and
  Fudan University, Shanghai, China
}

\authorrunning{Qi'ao Xu et al.}

\maketitle

\begin{abstract}
A core capability towards general embodied intelligence lies in localizing task-relevant objects from an egocentric perspective, formulated as Spatio-Temporal Video Grounding (STVG). Despite recent progress, existing STVG studies remain largely confined to object-centric and descriptive instructions, neglecting the task-oriented reasoning that is crucial for embodied agents to accomplish goal-directed interactions. To bridge this gap, we introduce \textbf{ToG-Bench}, the first task-oriented spatio-temporal video grounding benchmark for egocentric videos. ToG-Bench is characterized by three key features: (1) \textbf{Task-oriented Grounding}, which requires identifying and localizing objects based on intended tasks rather than straightforward descriptions; (2) \textbf{Explicit–Implicit Dual Grounding}, where target objects can be either explicitly mentioned or implicitly inferred by contextual reasoning; (3) \textbf{One-to-Many Grounding}, where a single instruction may correspond to multiple objects involved in task execution. Built upon videos sourced from ScanNet, ToG-Bench comprises 100 annotated clips with 2,704 task-oriented grounding instructions, constructed via a semi-automated pipeline that combines foundation model annotation and human refinement. In addition, we introduce a set of task-level evaluation metrics tailored for multi-object and explicit–implicit object grounding, and systematically benchmark seven state-of-the-art MLLMs. Extensive experiments reveal the intrinsic challenges of task-oriented STVG and substantial performance gaps across explicit–implicit and multi-object grounding, highlighting the difficulty of bridging perception and interaction in embodied scenarios. Data and code will be released.
\end{abstract}    
\section{Introduction}
\label{sec:intro}

Unlike traditional AI systems that operate solely in virtual environments, embodied agents leverage egocentric video as their primary sensory input to perceive and interact with their surroundings~\cite{duan2022survey,plizzari2024outlook,fu2024objectrelator,li2025clivis, long2025survey}. To enable such agents to effectively understand human instructions and interact with their environment, Spatio-Temporal Video Grounding (STVG)~\cite{zhang2020does,yang2022tubedetr,wasim2024videogrounding} emerges as a fundamental capability: it requires models to accurately localize the instruction-referred objects along both the spatial and temporal dimensions of a video.

\begin{figure}[t]
    \centering
    \includegraphics[width=0.75\linewidth]{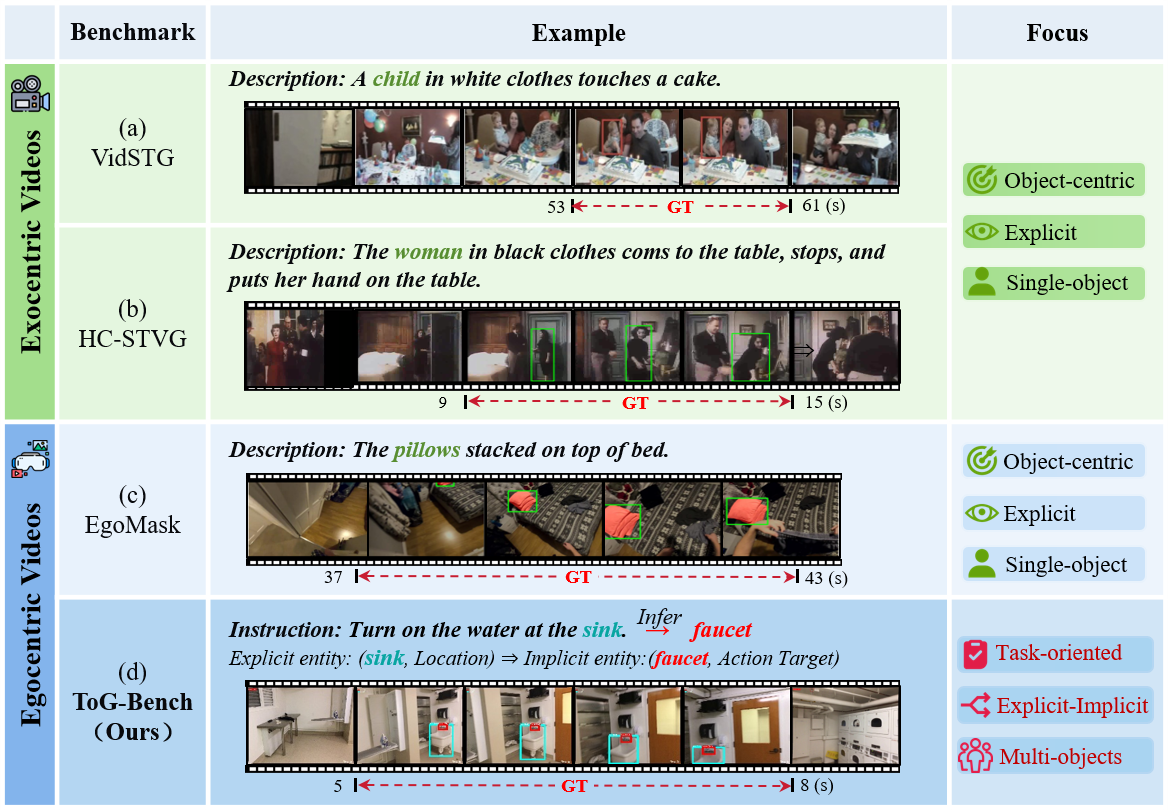}
  \vspace{-2mm}
    \caption{Illustration and comparison of existing STVG benchmarks and our ToG-Bench. While prior datasets (a–c) focus on object-centric, explicit, single-object grounding in either exocentric or egocentric videos, ToG-Bench (d) introduces a task-driven paradigm that supports \textbf{task-oriented} instructions, \textbf{explicit-implicit dual grounding}, and \textbf{multi-object grounding}, enabling robust evaluation of embodied agents.}
    \label{fig:motivation}
    \vspace{-1.5em}
\end{figure}

Despite remarkable progress in STVG driven by the emergence of dedicated benchmarks and powerful multimodal large language models (MLLMs), existing studies remain limited in its applicability to real-world embodied agents. On one hand, early benchmarks such as STPR~\cite{yamaguchi2017spatio}, VidSTG~\cite{zhang2020does}, and HC-STVG~\cite{tang2021human} are constructed from exocentric videos, where the camera viewpoint is detached from the agent’s embodiment. Such perspective neglect egocentric perceptual dynamics, such as attention shift, and agent-object interaction that are essential for instructions grounding~\cite{grauman2024ego,luo2024put}.
On the other hand, although recent efforts, such as RefEgo~\cite{Kurita_2023_ICCV} and EgoMask~\cite{liang2025fine}, have begun to explore STVG in egocentric videos, they still formulate the task in an object-centric and descriptive manner. As shown in Fig.~\ref{fig:motivation}, their instructions typically focus on identifying or describing visible entities (e.g., \textit{The pillows stacked on top of bed}) rather than conveying some intent (e.g., \textit{Turn on the water at the sink}), which prevents current benchmarks from supporting goal-directed embodied tasks that require grounding objects by their functional roles in a specific action. 

To fill the aforementioned gap, we formally define a new task, Task-oriented Spatio-Temporal Video Grounding (T-STVG), which extends traditional STVG toward goal-driven grounding of task-relevant objects. The objective is to identify and localize all objects involved in performing a given task within egocentric video. T-STVG presents three distinctive challenges that go beyond the scope of prior research. Specifically, it involves (1) \textbf{Task-driven reasoning}: the agent must infer task-relevant objects from high-level goals rather than relying on direct object description (e.g., \textit{make coffee} requires grounding a \textit{coffee machine}); (2) \textbf{Coexistence of explicit and implicit objects}: target entities may be explicitly mentioned in the instruction or implicitly inferred through contextual reasoning (e.g., \textit{turn on the water at the sink} involves an explicit object \textit{sink}, as well as an implicit one, \textit{faucet}, which is not mentioned in the instruction but is essential for completing the task); (3) \textbf{Multi-object grounding}: successful task execution often depends on coordinated localization of multiple interrelated objects (e.g., \textit{prepare the whiteboard for the next brainstorming session} involves \textit{whiteboard}, \textit{marker}, and \textit{eraser}).


Following these principles, we construct ToG-Bench, a novel semi-automatically built benchmark through the collaboration between vision-language foundation models and human refinement. ToG-Bench contains 100 video clips (with an average length of 87.88 seconds) sourced from ScanNet~\cite{dai2017scannet}, annotated with 2,704 task-oriented instructions and 4,194 object instances. Each object is annotated with second-level precise temporal boundaries and frame-level bounding boxes. In addition, to systematically evaluate the performance of T-STVG, we propose a suite of task-level metrics that collectively assess object recognition accuracy (i.e., the ability to correctly identify task-related objects) and spatio-temporal precision (i.e., the ability to localize them precisely in both spatial and temporal dimensions). We benchmark {eight} state-of-the-art MLLMs on ToG-Bench, revealing that while models perform reasonably well on explicit single-object cases, their performance drops sharply on implicit reasoning and multi-object grounding. These results indicate that ToG-Bench poses substantial challenges, providing a new benchmark for studying task-oriented grounding and promoting progress toward more capable embodied agents.

Our main contributions are summarized as follows:
\begin{itemize}
 \renewcommand{\labelitemi}{$\bullet$}
    \item We introduce Task-oriented Spatio-Temporal Video Grounding (T-STVG), a novel task that advances traditional STVG toward task-driven grounding in egocentric perspectives, and establish a dedicated benchmark, ToG-Bench, to support systematic evaluation.
    \item We design task-level evaluation metrics that jointly assess recognition accuracy and spatio-temporal localization precision for task-relevant objects.
    \item Comprehensive experiments with state-of-the-art MLLMs reveal the intrinsic difficulty of T-STVG, especially in implicit reasoning and multi-object grounding, and highlight its potential in embodied AI.
\end{itemize}


\section{Related Work}
\label{sec:related_work}

\subsection{Spatio-Temporal Video Grounding}
Spatio-Temporal Video Grounding (STVG) aims to localize both the temporal span and spatial region within a video according to a language query. 
This task stems from image-level Visual Grounding (VG), which has evolved from descriptive localization (e.g., RefCOCO/+/g~\cite{kazemzadeh2014referitgame,yu2016modeling,mao2016generation}) to intent-based spatial reasoning (e.g., IntentionVG~\cite{wang2024beyond}, EgoIntention~\cite{sun2025visual}). However, these VG methods remain confined to static frames and lack temporal dimensions.
In the video domain, early benchmarks such as Charades-STA~\cite{gao2017tall} and ActivityNet-Captions~\cite{krishna2017dense} primarily focus on exocentric videos with descriptive queries, providing only temporal annotations without spatial grounding. Subsequent work extended grounding to include spatial localization, including STPR~\cite{yamaguchi2017spatio}, Vid-Sentence~\cite{chen2019weakly}, VidSTG~\cite{zhang2020does}, HC-STVG~\cite{tang2021human}, DVD-ST~\cite{ji2024described}, and BOSTVG~\cite{yao2025omnistvg}.
Recently, egocentric benchmarks such as Ego4D~\cite{grauman2022ego4d}, RefEgo~\cite{Kurita_2023_ICCV}, and EgoMask~\cite{liang2025fine} have advanced STVG in first-person settings by enabling temporal and spatial grounding in real-world videos.
Despite these advancements, current STVG efforts remain predominantly object-centric, focusing on explicit one-to-one alignments between text and visual targets. They often struggle with task-oriented instructions that require reasoning about implicit targets and multi-object interactions over time. To address this limitation, we introduce ToG-Bench, the first task-oriented STVG benchmark for egocentric videos. ToG-Bench shifts the paradigm from simple object description to high-level reasoning about task-relevant objects in dynamic environments.

\subsection{Multimodal Large Language Models}
Current Multimodal Large Language Models (MLLMs) have demonstrated remarkable advancements in visual-language tasks like captioning, question answering, and reasoning.
Early MLLMs focus on unimodal localization. For temporal grounding, VTimeLLM~\cite{huang2024vtimellm} refines timestamps via boundary-aware training, while TRACE~\cite{guo2024trace} uses causal modeling. For spatial grounding, GroundingGPT~\cite{li2024groundinggpt} enhances alignment through language-guided fusion. Recent efforts like SpaceVLLM~\cite{wang2025spacevllm} and LLaVA-ST~\cite{li2025llava} enable joint spatio-temporal prediction, yet they remain evaluated on descriptive queries, not task-driven instructions.
While general-purpose MLLMs such as GPT-5~\cite{achiam2023gpt,openai2025gpt5}, Gemini 2.5 Pro~\cite{comanici2025gemini}, and Qwen3VL~\cite{qwen3technicalreport} demonstrate broad video understanding, their ability to localize task-relevant objects under implicit references or multi-object coordination in egocentric videos remains largely untested, due to the lack of suitable benchmarks.
ToG-Bench fills this gap by introducing the first benchmark designed to assess MLLMs on task-oriented spatio-temporal grounding in egocentric videos, revealing critical limitations and guiding future development toward embodied intelligence.

\section{ToG-Bench}
\label{sec:dataset}


\subsection{Task Formulation}
We formally define Task-oriented Spatial-temporal Video Grounding (T-STVG) as a mapping function $\mathcal{F}: (G, \mathcal{V}) \to \{(O_i, \mathcal{T}_i)\}_{i=1}^n$. Given a high-level goal $G$ expressed in natural language and an egocentric video sequence $\mathcal{V}$, the model is required to autonomously infer $n$ task-relevant objects $\{O_i\}_{i=1}^n$ and predict their corresponding dense spatio-temporal trajectories $\mathcal{T}_i = \{ (b_{i,t}, t) \mid t \in [t_{start}, t_{end}] \}$, where $b_{i,t}$ denotes the bounding box of object $O_i$ at frame $t$.

Unlike conventional STVG benchmarks that adopt a \emph{descriptive grounding paradigm} to identify objects by visual attributes (\emph{e.g.}, color, shape, or state), T-STVG redefines grounding as a \emph{task-oriented paradigm}. This approach aligns visual grounding with action intent, requiring the model to perform \emph{functional reasoning} over the environment. To implement this paradigm, we constructed ToG-Bench following three key principles:

\begin{itemize}
\label{subsec:toi}
 \renewcommand{\labelitemi}{$\bullet$}
    \item Task-Oriented Instruction: T-STVG formulates $G$ as goal-directed actions, emphasizing interaction with the environment (\emph{e.g.}, ``\emph{I'm a bit thirsty and want some water.}''). The objects $O_i$ are defined by the tools or items available in the agent’s current view associated with the task, such as a ``\emph{water bottle}'' used for drinking.
    \item Explicit-Implicit Dual Grounding: The target set is defined as $\{O_i\} = \mathcal{O}_{exp} \cup \mathcal{O}_{imp}$. While $\mathcal{O}_{exp}$ comprises entities explicitly mentioned (\emph{e.g.}, ``\emph{sink}'' in ``\emph{wash hands at the sink}''), $\mathcal{O}_{imp}$ denotes implicit objects (\emph{e.g.}, ``\emph{faucet}'') that must be inferred through contextual reasoning and commonsense knowledge.
    \item Multi-object Grounding: ToG-Bench supports one-to-many grounding ($n \geq 1$), reflecting the complexity of real-world embodied interactions. For instance, ``\emph{Work on your computer.}'' may require the simultaneous localization of a \emph{computer}, \emph{mouse}, and \emph{keyboard}. 
\end{itemize}

These principles are mutually reinforcing, shifting the grounding criterion from perceptual similarity to functional relevance. The task-oriented formulation naturally induces implicit references and multi-object dependencies, demanding higher-level functional understanding beyond low-level visual perception to bridge the gap between perception and intention.

\subsection{Semi-automated Annotation Pipeline}

\begin{figure}[t]
    \centering
    \includegraphics[width=0.9\linewidth]{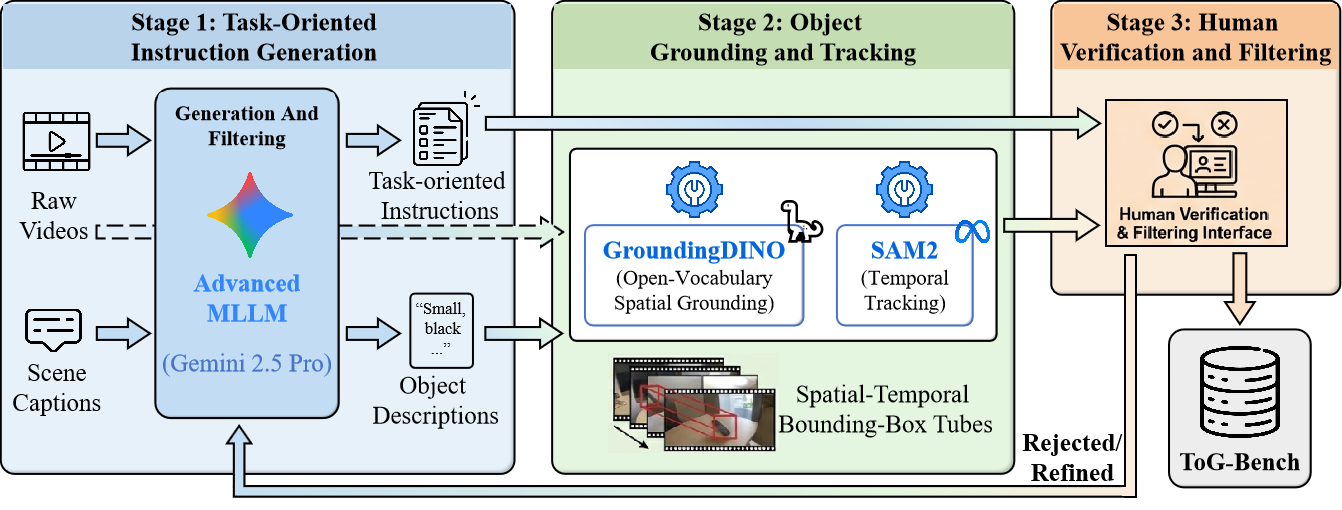}
    \caption{Semi-automated Annotation Pipeline. The process consists of three top-down stages: (1) generating task-oriented instructions and corresponding object descriptions with MLLM, (2) grounding and tracking task-relevant objects via Grounding-DINO and SAM2, and (3) performing human verification and filtering.}
    \label{fig:pipeline}
    \vspace{-1em}
\end{figure}

We develop a semi-automated annotation pipeline that combines vision language foundation models with human verification to efficiently construct the ToG-Bench dataset. As illustrated in Fig.~\ref{fig:pipeline}, the pipeline consists of three stages: (1) \emph{Task-Oriented Instruction Generation.} We feed the raw video clips from ScanNet~\cite{dai2017scannet} test-set along with their accompanying scene captions into an advanced MLLM (Gemini 2.5 Pro~\cite{comanici2025gemini}) to generate a set of task-oriented instructions and detailed descriptions of the objects involved in each task, covering both explicit and implicit forms. (2) \emph{Object Grounding and Tracking.} Based on the object descriptions, we apply Grounding-DINO~\cite{liu2024grounding} and SAM2~\cite{ravi2024sam} in tandem to perform open-vocabulary spatial grounding and temporal tracking, resulting in spatio-temporal bounding-box tubes for task-relevant objects. (3) \emph{Human Verification and Filtering.} Since foundation models may hallucinate or generate imprecise boundaries, all generated instructions and object annotations undergo rigorous manual validation.

A key characteristic of our annotation pipeline lies in its top-down design philosophy. In contrast to conventional STVG benchmarks, which follow a bottom-up procedure (detecting and tracking all objects first, then associating descriptive instructions), our pipeline starts with instruction generation and only grounds the objects relevant to those instructions. This top-down strategy aligns with the task-oriented nature of T-STVG, as many scene objects (\emph{e.g.}, wall or other static background items) are not actionable for task execution. Detecting and tracking every object beforehand would introduce substantial noise and annotation overhead, whereas our approach ensures efficiency and semantic focus. From 8,334 initial candidates, our multi-stage filtering process yields 2,704 high-quality instructions, representing a 32.4\% retention rate. In the following, we will elaborate on each stage of the annotation pipeline in detail.

\paragraph{Task-Oriented Instruction Generation.}
We begin by generating a structured caption for each video with Gemini 2.5 Pro, the caption captures the object attributes and their relationships in the scene. Then the generated captions, together with the video frames, are integrated as contextual cues to facilitate the generation of task-oriented instructions. To ensure the quality, we adopt two prompt strategies to generate 4,010 explicit and 4,324 implicit candidates. All generated instructions are subsequently screened through an automatic verification stage to ensure: 
(1) \emph{Visibility}: all referred objects are present and clearly visible in the video;
(2) \emph{Uniqueness}: each target object is unambiguously identifiable given attributes and spatial context;
(3) \emph{Feasibility}: unsafe or impractical actions are filtered out.
This stage retained 58.3\% of explicit and 57.9\% of implicit tasks (4,842 total).
Detailed prompts can be found in the Appendix.

\paragraph{Object Grounding and Tracking.}
For each validated instruction, we extract target object and contextual cues such as attributes, states, and relations to synthesize detailed object descriptions. Based on these descriptions, we apply Grounding-DINO to intermittently capture target objects in keyframes across the video. These detected bounding boxes then serve as visual prompts for SAM2, which performs bi-directional tracking to propagate the boxes forward and backward, producing complete spatio-temporal grounding tubes (\emph{i.e.}, a sequence of bounding boxes aligned with the task execution timeline).
To balance the coverage and computational cost, we perform object tracking at a fixed frame rate of 1 fps.

\paragraph{Human Verification and Filtering.}
We visualize all spatio-temporal grounding tubes and conduct two rounds of human validation:
(1) \emph{First-pass filtering}: remove trajectories affected by occlusion, appearance changes, or tracking drift, especially critical for implicitly referred objects. 79.3\% (3,840) of the 4,842 candidates were retained after purging mismatches.
(2) \emph{Second-pass verification}: verify temporal boundaries align with task execution phases and ensure consistency in multi-object interactions. This final refinement yielded 2,704 instructions (70.4\% retention), with all misaligned or low-quality annotations strictly filtered out.
Through this semi-automated pipeline, we construct ToG-Bench, a high-quality benchmark for task-oriented spatio-temporal video grounding, with precise, diverse, and reliably grounded annotations.


\begin{figure}[t]
    \centering
    \includegraphics[width=0.98\linewidth]{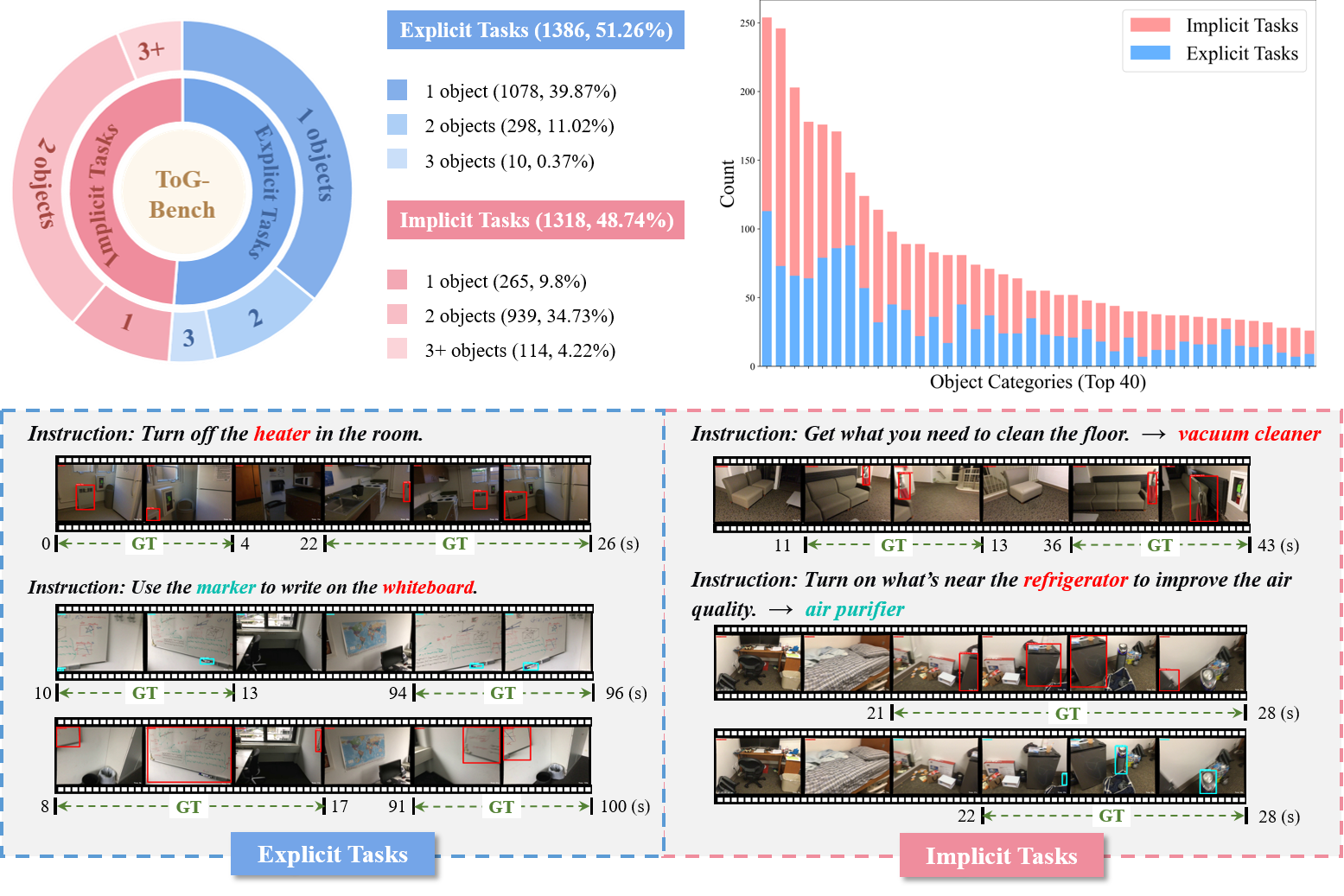}
    \vspace{-2mm}
    \caption{Dataset characteristics of ToG-Bench. \textbf{Top-left}: Task distribution by type (explicit vs. implicit) and object count; \textbf{Top-right}: Object category frequency (top 40 categories); \textbf{Bottom}: Example grounding tubes for explicit (blue) and implicit (pink) tasks, highlighting contextual inference and multi-object grounding.}
    \label{fig:dataset}
    \vspace{-1em}
\end{figure}

\subsection{Dataset Statistics}

\begin{figure}[t]
    \centering
    \begin{minipage}{0.52\linewidth}
        \centering
        \includegraphics[width=\linewidth]{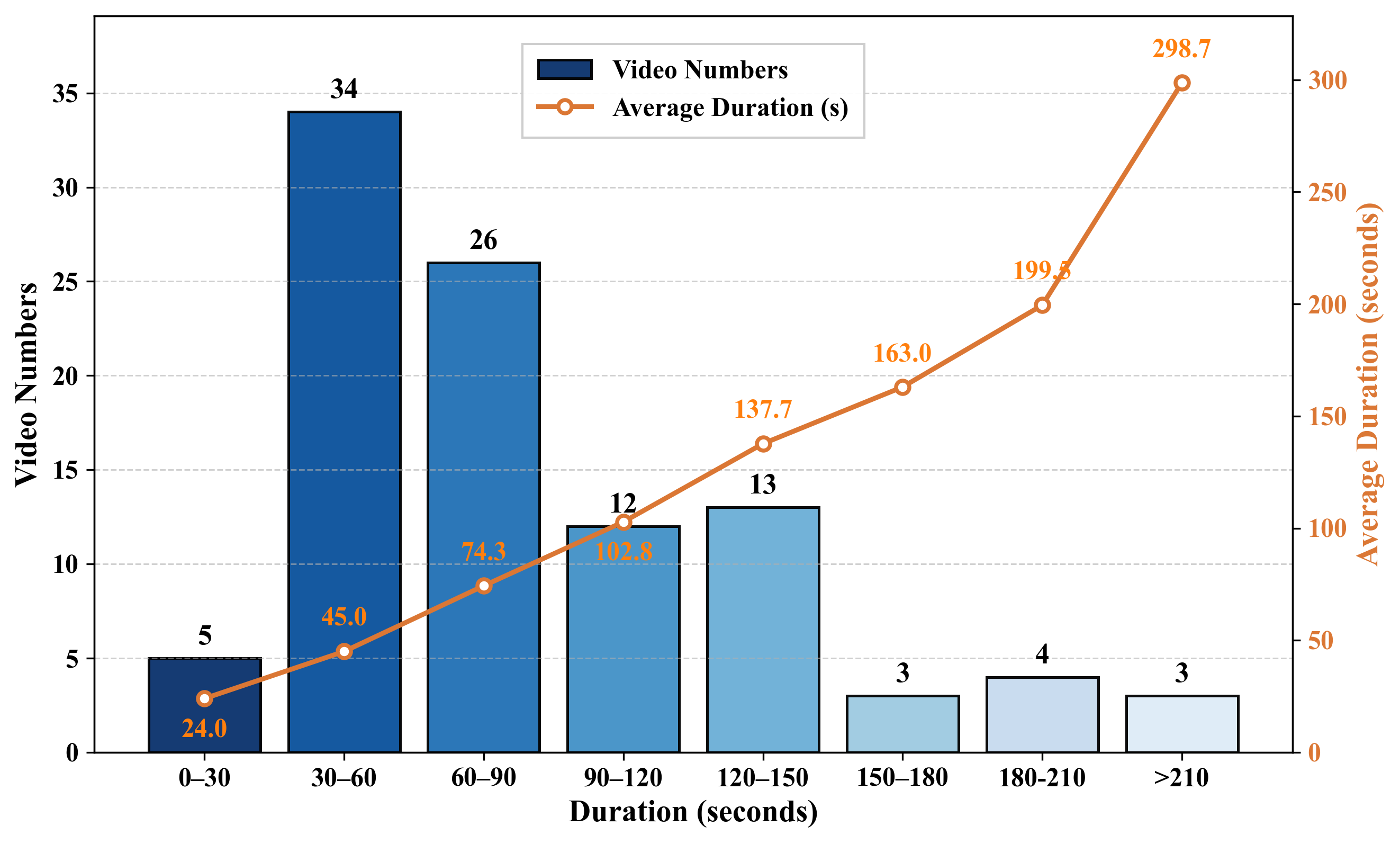}
    \end{minipage}
    \hspace{0.02\linewidth}
    \begin{minipage}{0.36\linewidth}
        \centering
        \scriptsize
        \renewcommand{\arraystretch}{1.1}
        \begin{tabular}{c|cccc}
            \toprule
            \textbf{Interval (s)} & \textbf{Video} & \textbf{Task} & \textbf{Exp.} & \textbf{Imp.} \\
            \midrule
            0--30    & 5  & 136 & 62  & 74  \\
            30--60   & 34 & 820 & 418 & 402 \\
            60--90   & 26 & 710 & 357 & 353 \\
            90--120  & 12 & 346 & 180 & 166 \\
            120--150 & 13 & 381 & 196 & 185 \\
            150--180 & 3  & 85  & 47  & 38  \\
            180--210 & 4  & 143 & 75  & 68  \\
            $>$210   & 3  & 83  & 51  & 32  \\
            \midrule
            \textbf{Total} & \textbf{100} & \textbf{2,704} & \textbf{1,386} & \textbf{1,318} \\
            \bottomrule
        \end{tabular}
    \end{minipage}
    \vspace{-2mm}
    \caption{Video duration and task distribution. {Left}: Video count per interval and average duration. {Right}: Detailed breakdown of task counts, including explicit (Exp.) and implicit (Imp.) types.}
    \label{fig:video_duration_combined}
    \vspace{-1em}
\end{figure}

In total, ToG-Bench consists of 100 egocentric videos simulated from first-person trajectories in ScanNet indoor environments, 2,704 task-oriented spatio-temporal grounding instructions, and 4,194 annotated object instances. Figures~\ref{fig:dataset} and \ref{fig:video_duration_combined} illustrate the distributions across instructions, objects, and video durations.

\paragraph{Instruction Distribution.}
As illustrated in chart of Fig.~\ref{fig:dataset} (top-left), ToG-Bench comprises 2,704 instructions, corresponding to an average of 27.0 tasks per video. Among these, 1,386 (51.26\%) are explicit tasks that directly refer the target objects, while 1,318 (48.74\%) are implicit tasks that require contextual reasoning to identify the relevant entities. This near-balanced distribution ensures that the benchmark equally challenges models on both direct perception and high-level reasoning capabilities. Regarding one-to-many grounding, 1,361 tasks (50.33\%) involve two or more objects, including 124 tasks (4.6\%) that reference three or more objects. On average, implicit tasks ground 1.97 objects per instruction, compared to 1.14 objects per explicit task, underscoring the inherently higher semantic complexity and reasoning demands of implicit grounding. The bottom panel of Fig.~\ref{fig:dataset} provides examples of explicit and implicit tasks with time-stamped frames and ground-truth (GT) tubes overlaid, illustrating the task-driven characteristics of our proposed benchmark.



\paragraph{Object Distribution.} 
{As shown in the chart of Fig.~\ref{fig:dataset} (top-right)}, ToG-Bench includes 4,194 object instances (mean: 41.94 per video), spanning 177 distinct functional object categories derived through semantic clustering. These categories are widely distributed across explicit and implicit tasks, with 145 categories appearing in explicit tasks and 164 in implicit tasks. This indicates that implicit grounding engages a similarly diverse set of functionally relevant objects as explicit grounding, rather than a narrow subset. The diversity of objects across tasks further highlights the challenge posed by ToG-Bench.

\paragraph{Video duration and task distribution.} The total duration of ToG-Bench is 8,788 seconds, with individual clips ranging from 13 to 347 seconds (mean: 87.88 s). As shown in Fig.~\ref{fig:video_duration_combined}, the 2,704 total instructions are distributed across varying time horizons to cover both atomic actions and procedural activities. Most tasks fall in the 30--150 s range, which accounts for 2,257 instructions across 85 videos. This temporal diversity requires models to perform consistent spatio-temporal reasoning across varying time horizons, from fine-grained actions to long-term task dependencies.

Together, these characteristics make ToG-Bench a comprehensive and challenging benchmark for evaluating task-oriented spatio-temporal grounding in embodied video understanding.


\section{Experiments}
\label{sec:experiment}

\subsection{Experimental Setup}

\paragraph{Evaluated Models.}
We evaluate two representative categories of multimodal large language models (MLLMs) to assess their zero-shot capabilities on task-oriented spatio-temporal grounding (T-STVG) in egocentric videos:
(1) Proprietary MLLMs: GPT-5 \cite{openai2025gpt5} and Gemini 2.5 Pro~\cite{comanici2025gemini}. (2) Open-source general-purpose MLLMs: Qwen3-VL (4B, 8B) \cite{qwen3technicalreport}, Qwen2.5-VL-7B \cite{bai2025qwen2}, VideoLLaMA3-7B \cite{zhang2025videollama}, InternVL3-8B \cite{zhu2025internvl3}, and InternVL3.5-8B \cite{wang2025internvl3}. 
In addition, we include Grounded-SAM2 \cite{ravi2024sam} as a specialized grounding baseline to provide a performance reference. As Grounded-SAM2 lacks high-level reasoning capabilities, we bypass the instruction inference stage by directly providing it with the specific target object names for each task.

\paragraph{Implementation Details.}
\label{par:implementation_deatils}
All models are assessed under a zero-shot, single-pass inference protocol to evaluate their inherent grounding capabilities without any task-specific fine-tuning. Videos are sampled at 0.25 frames per second, optimizing the trade-off between temporal coverage and computational efficiency, with no restriction on the number of frames to ensure complete contextual understanding. Open-source models are run on NVIDIA A100 40GB GPUs, whereas proprietary models are evaluated through their official APIs.

\subsection{Evaluation Metrics}

\begin{figure}[b!]
\centering
\begin{minipage}{0.90\columnwidth}\vspace{0mm}   
\centering
\begin{tcolorbox} 
    \raggedright
    \small
     \hspace{-6mm}
    
    Please carefully analyze the video frames and the given task instruction... Based on your visual understanding of the scene, please first \textcolor{red}{identify the object category} ... Then provide \textcolor{red}{temporal intervals} and \textcolor{red}{spatial coordinates} of ... 
\end{tcolorbox}
\vspace{-2mm}
\caption{A concise demonstration of the system prompt utilized for inference.}
    \label{tab:eval_prompt}
\end{minipage}
\end{figure}

Considering the \textit{``recognition-then-grounding''} logic inherent in task-oriented instructions depicted in Sec.~\ref{subsec:toi}, we implement a decoupled evaluation paradigm. The system prompt employed for inference is shown in Fig.~\ref{tab:eval_prompt}, where we first prompt the MLLM to infer the category of the referred object, followed by predicting the time intervals or spatial coordinates. For metric evaluation, we first assess the MLLM’s \textbf{reasoning capabilities} in uncovering the target object embedded in task-oriented user queries within the embodied context. Following this, with the recognized target objects serving as guidance, we further investigate the MLLM’s \textbf{grounding capabilities} in both the spatial and temporal dimensions.
(1) For reasoning capability evaluation, we employ accuracy (Acc) as the metric. Concretely, with the predicted object category, we match it with the ground-truth referred object by calculating the cosine similarity between their text embeddings extracted by \texttt{all-mpnet-base-v2}~\cite{lo-wang-2020-s2orc}. We also group objects by task types (i.e, explicit and implicit), and calculate their recognition accuracy, denoted as EAcc and IAcc, respectively.
(2) For grounding capability evaluation, we calculate recall (R) for temporal grounding, average precision (AP) for spatial grounding, as well as mIoU for both tasks. Concretely, we apply different $\tau$ ($\tau \in \{0.3, 0.5, 0.7\}$) as thresholds when calculating the recall and average precision following prior works~\cite{cheng2025v,zhang2020does,yao2025omnistvg}, which are denoted as R1@$\tau$ and AP@$\tau$, respectively. 
Furthermore, to comprehensively evaluate model performances, we calculate the above metrics from both object-level and task-level perspectives, and the corresponding metrics are denoted as O-Acc, O-R1@$\tau$, O-AP@$\tau$, and T-Acc, T-R1@$\tau$, T-AP@$\tau$, respectively. 
While object-level metrics treat each entity independently, task-level evaluation is stricter as it assesses the fulfillment of the entire instruction. A task is considered successful only if all required objects are correctly recognized and grounded, since any failure to identify even a single target results in a zero score for the entire task.

\subsection{Main Results}

\begin{table}[t]
  \centering
  \caption{Task-level performance comparison of MLLMs on ToG-Bench. The best result is in \textbf{bold}, and the second is \underline{underlined}. All values are reported in percentage (\%).}
  \vspace{-2mm}
  \resizebox{\linewidth}{!}{
    \begin{tabular}{l|c|cccc|cccc}
      \toprule
      \multirow{2}{*}{\textbf{Models}} & \multirow{2}{*}{\textbf{T-Acc}} & \multicolumn{4}{c|}{\textbf{Task-level Temporal Video Grounding}}  & \multicolumn{4}{c}{\textbf{Task-level Spatial Video Grounding}} \\
      &  & {T-m\_tIoU} & {T-R1@0.3} & {T-R1@0.5} & {T-R1@0.7} & {T-m\_vIoU} & {T-AP@0.3} & {T-AP@0.5} & {T-AP@0.7} \\
      \midrule
      \multicolumn{10}{c}{\textit{Grounding Baselines}} \\
      \midrule
      Grounded-SAM2 \cite{ravi2024sam} & - & 26.21 & 24.11 & 8.65 & 3.55 & \textbf{22.06} & \textbf{16.72} & \underline{3.88} & \underline{1.55} \\
      \midrule
      \multicolumn{10}{c}{\textit{Proprietary MLLMs}} \\
      \midrule
      GPT-5~\cite{openai2025gpt5} & \textbf{89.42} & \underline{37.06} & \underline{43.64} & \underline{27.33} & \underline{13.17} & 10.22 & 6.21 & 1.89 & 0.18 \\
      Gemini 2.5 Pro~\cite{comanici2025gemini} & \underline{80.14} & \textbf{40.52} & \textbf{49.37} & \textbf{34.17} & \textbf{19.12} & \underline{15.42} & \underline{15.35} & \textbf{5.88} & \textbf{1.59} \\
      \midrule
      \multicolumn{10}{c}{\textit{Open-source MLLMs}} \\
      \midrule
      Qwen2.5-VL-7B~\cite{bai2025qwen2} & 54.70 & 7.43 & 9.62 & 2.44 & 0.30 & 0.47 & 0.00 & 0.00 & 0.00 \\
      Qwen3-VL-4B~\cite{qwen3technicalreport} & 49.67 & 1.18 & 1.11 & 0.48 & 0.07 & 0.81 & 0.04 & 0.04 & 0.00 \\
      Qwen3-VL-8B~\cite{qwen3technicalreport} & 65.13 & 16.20 & 19.82 & 8.39 & 2.37 & 1.48 & 0.18 & 0.07 & 0.00 \\
      VideoLLaMA3-7B~\cite{zhang2025videollama} & 28.55 & 2.80 & 2.92 & 1.18 & 0.44 & 0.58 & 0.00 & 0.00 & 0.00 \\
      InternVL3-8B~\cite{zhu2025internvl3} & 61.39  & 7.98 & 7.14 & 2.59 & 0.52 & 0.60 & 0.00 & 0.00 & 0.00 \\
      InternVL3.5-8B~\cite{wang2025internvl3} & 52.07 & 8.70 & 10.32 & 4.22 & 1.15 & 1.02 & 0.00 & 0.00 & 0.00 \\
      \bottomrule
    \end{tabular}
  }
  \label{tab:mllm_performance}
\vspace{-1em}
\end{table}

\paragraph{Overall Performance Comparison.}
Table \ref{tab:mllm_performance} presents the task-level performance of leading MLLMs and specialized grounding models on ToG-Bench, highlighting the significant challenge of task-oriented grounding in egocentric videos. 
(1) Performance Hierarchy: A consistent difficulty gap exists where task accuracy (T-Acc) is highest, followed by temporal and then spatial grounding. For example, Gemini 2.5 Pro achieves 80.14\% T-Acc but only 15.42\% T-m\_vIoU, indicating that high-level intent recognition is significantly easier than precise spatial-temporal localization in MLLMs.
(2) Proprietary MLLM Trade-offs: While GPT-5 leads in task reasoning (89.42\% T-Acc), Gemini 2.5 Pro demonstrates superior spatial-temporal alignment, outperforming GPT-5 in both temporal (40.52\%) and spatial (15.42\%) dimensions. This highlights a potential decoupling between high-level task inference and fine-grained spatial-temporal alignment.
(3) Specialist vs. Generalist: To provide a spatial reference, we evaluate Grounded-SAM2 by providing it with oracle object names. Despite this privileged setting, it achieves the highest spatial score (22.06\% T-m\_vIoU), highlighting MLLMs' limitations in fine-grained localization. However, its temporal scores (26.21\% T-m\_tIoU) remain inferior to Gemini 2.5 Pro and GPT-5, proving that spatial precision cannot substitute for the task-level temporal reasoning essential to T-STVG.
(4) Open-source Bottleneck: Open-source models like Qwen3-VL-8B struggle significantly, with spatial metrics dropping to 1.48\%. This systematic failure underscores the difficulty of jointly resolving the ``what-when-where'' triplet under the rapid viewpoint shifts and occlusions typical of egocentric embodied scenarios.


\begin{table}[t]
  \centering
  \caption{Object-level performance comparison of MLLMs on ToG-Bench.}
  \vspace{-2mm}
  \resizebox{\linewidth}{!}{%
    \begin{tabular}{l|ccc|cccc|cccc} 
      \toprule
      \multirow{2}{*}{\textbf{Models}} & \multirow{2}{*}{\textbf{O-Acc}} & \multirow{2}{*}{\textbf{EAcc}} & \multirow{2}{*}{\textbf{IAcc}} & \multicolumn{4}{c|}{\textbf{Temporal Video Grounding}} & \multicolumn{4}{c}{\textbf{Spatial Video Grounding}} \\
      & &  &  & {O-m\_tIoU} & {O-R1@0.3} & {O-R1@0.5} & {O-R1@0.7} & {O-m\_vIoU} & {O-AP@0.3} & {O-AP@0.5} & {O-AP@0.7} \\
      \midrule
      \multicolumn{12}{c}{\textit{Grounding Baselines}} \\
      \midrule
      Grounded-SAM2 \cite{ravi2024sam} & - & - & - & 26.59 & 34.93 & 15.43 & 6.37 & \textbf{22.45} & \textbf{27.63} & \textbf{8.23} & \textbf{2.58} \\
      \midrule
      \multicolumn{12}{c}{\textit{Proprietary MLLMs}} \\
      \midrule
      GPT-5~\cite{openai2025gpt5}  & \textbf{92.82} & \textbf{95.60} & \textbf{86.98} & \underline{37.84} & \underline{52.74} & \underline{34.86} & \underline{18.12} & 10.22 & 8.66 & 2.12 & 0.14 \\
      Gemini 2.5 Pro~\cite{comanici2025gemini} & \underline{84.55} & \underline{87.40} & \underline{78.55} & \textbf{41.55} & \textbf{57.39} & \textbf{41.82} & \textbf{24.68} & \underline{15.76} & \underline{21.01} & \underline{7.92} & \underline{2.15} \\
      \midrule
      \multicolumn{12}{c}{\textit{Open-source MLLMs}} \\
      \midrule
      Qwen2.5-VL-7B~\cite{bai2025qwen2} & 66.09 & 68.61 & 60.80 & 8.77 & 12.49 & 3.03 & 0.48 & 0.52 & 0.02 & 0.00 & 0.00 \\
      Qwen3-VL-4B~\cite{qwen3technicalreport} & 61.59 & 60.34 & 64.20 & 1.45 & 1.60 & 0.74 & 0.26 & 0.94 & 0.07 & 0.05 & 0.00 \\
      Qwen3-VL-8B~\cite{qwen3technicalreport} & 74.42 & 76.50 & 70.04 & 17.94 & 25.70 & 11.71 & 3.41 & 1.60 & 0.26 & 0.07 & 0.00 \\
      VideoLLaMA3-7B~\cite{zhang2025videollama}  & 32.50 & 36.14 & 24.85 & 3.13 & 4.01 & 1.67 & 0.62 & 0.66 & 0.12 & 0.02 & 0.00 \\
      InternVL3-8B~\cite{zhu2025internvl3}  & 68.88 & 71.75 & 62.87 & 9.25 & 10.47 & 3.74 & 0.88 & 0.63 & 0.02 & 0.02 & 0.00 \\
      InternVL3.5-8B~\cite{wang2025internvl3} & 62.71 & 63.27 & 61.54 & 10.17 & 14.02 & 5.65 & 1.55 & 1.11 & 0.02 & 0.00 & 0.00 \\
      \bottomrule
    \end{tabular}
  }
  \label{tab:object_level_performance}
\vspace{-1em}
\end{table}

\paragraph{Object-level Performance Comparison.}
Table \ref{tab:object_level_performance} reports the object-level performance on ToG-Bench, offering a more fine-grained assessment of individual entity grounding. 
(1) Implicit Reasoning Challenge: While semantic accuracy is high, performance drops significantly on implicit tasks. For instance, GPT-5’s accuracy falls from 95.60\% (EAcc) to 86.98\% (IAcc). This gap, consistent across models, underscores the difficulty of inferring task-relevant objects through contextual reasoning when they are not explicitly mentioned.
(2) Task vs. Object Granularity: Task-level accuracy (T-Acc) is consistently lower than Object-level accuracy (O-Acc) (e.g., GPT-5: 89.42\% vs. 92.82\%; Qwen3-VL: 65.13\% vs. 74.42\%). This is because task-level evaluation imposes a stricter ``all-or-nothing'' criterion, where a single misidentified object results in task failure. Threshold-based metrics (R1/AP) show a sharper decline, confirming that grounding multiple interrelated objects simultaneously is a major bottleneck.
(3) Consistency of Model Behaviors: The object-level results mirror the model-specific trade-offs discussed previously. Specifically, the specialized Grounded-SAM2 maintains its lead in spatial precision (22.45\% O-m\_vIoU while remaining inferior in temporal grounding. Similarly, the catastrophic failure of open-source models in spatial metrics (e.g., Qwen3-VL-8B's 1.60\% O-m\_vIoU) is confirmed here. These results suggest that both the specialist-generalist gap and the open-source alignment bottleneck are systemic across different evaluation granularities, regardless of whether single or multiple objects are considered.
\begin{table}[t]
  \centering
  \caption{Task-level grounding performance on explicit and implicit tasks.}
\vspace{-2mm}
  \resizebox{\linewidth}{!}{
    \begin{tabular}{l|ccccc|ccccc}
      \toprule
      \multirow{2}{*}{\textbf{Models}} 
      & \multicolumn{5}{c|}{\textbf{Explicit Tasks}} 
      & \multicolumn{5}{c}{\textbf{Implicit Tasks}} \\
      & {T-EAcc} & {T-m\_EtIoU} & {T-ER1@0.5} & {T-m\_EvIoU} & {T-EAP@0.5}  & {T-IAcc} & {T-m\_ItIoU} & {T-IR1@0.5}  & {T-m\_IvIoU} & {T-IAP@0.5} \\
      \midrule
      \multicolumn{11}{c}{\textit{Grounding Baselines}} \\
      \midrule
      Grounded-SAM2 \cite{ravi2024sam} & - & 26.22 & 11.54 & \textbf{22.38} & 5.92 & - & 26.20 & 5.61 & \textbf{21.72} & \underline{1.75} \\
      \midrule
      \multicolumn{11}{c}{\textit{Proprietary MLLMs}} \\
      \midrule
      GPT-5~\cite{openai2025gpt5} & \textbf{98.20} & \underline{41.76} & \underline{34.78} & 12.36 & 3.10 & \textbf{80.20} & \underline{32.11} & \underline{19.50} & 7.98 & 0.61 \\
      Gemini 2.5 Pro~\cite{comanici2025gemini} & \underline{91.49}  & \textbf{47.67} & \textbf{44.88} & \underline{18.33} & \textbf{8.80} & \underline{68.21} & \textbf{33.01} & \textbf{22.91} & \underline{12.36} & \underline{2.81} \\

      \midrule
      \multicolumn{11}{c}{\textit{Open-source MLLMs}} \\
      \midrule
      Qwen2.5-VL-7B~\cite{bai2025qwen2} & 75.97 & 10.39 & 3.82 & 0.66 & 0.00 & 32.32 & 4.32 & 0.99 & 0.27 & 0.00 \\
      Qwen3-VL-4B~\cite{qwen3technicalreport} & 59.96 & 1.34 & 0.51 & 0.94 & 0.07 & 38.85 & 1.02 & 0.46 & 0.66 & 0.00 \\
      Qwen3-VL-8B~\cite{qwen3technicalreport}  & 75.97 & 19.52 &  12.41 & 1.80 & 0.14 & 53.72 & 12.72 & 4.17 & 1.15 & 0.00 \\
      VideoLLaMA3-7B~\cite{zhang2025videollama}  & 39.68 & 3.98 & 2.02 & 0.79 & 0.00 & 16.84 & 1.56 & 0.30 & 0.36 & 0.00 \\
      InternVL3-8B~\cite{zhu2025internvl3} & 81.60 & 10.42 & 4.26 & 0.82 & 0.00 & 40.14 & 5.41 & 0.83 & 0.37 & 0.00 \\
      InternVL3.5-8B~\cite{wang2025internvl3} & 64.72 & 10.62 & 5.99 & 1.33 & 0.00 & 38.77 & 6.68 & 2.35 & 0.69 & 0.00 \\
      \bottomrule
    \end{tabular}
  }
  \label{tab:explicit_implicit_core}
\end{table}

\begin{table}[t]
  \centering
  \caption{Task-level performance across tasks with varying numbers of target objects.}
  \vspace{-2mm}
  \resizebox{\linewidth}{!}{
    \begin{tabular}{l|ccc|ccc|ccc|ccc}
      \toprule
      \multirow{2}{*}{\textbf{Models}}  
      & \multicolumn{3}{c|}{\textbf{1 Object}} & \multicolumn{3}{c|}{\textbf{2 Objects}} & \multicolumn{3}{c|}{\textbf{3+ Objects}} & \multicolumn{3}{c}{\textbf{Avg.}} \\
      & {T-Acc} & {T-m\_tIoU} & {T-m\_vIoU} & {T-Acc} & {T-m\_tIoU} & {T-m\_vIoU} & {T-Acc} & {T-m\_tIoU} & {T-m\_vIoU} & {T-Acc} & {T-m\_tIoU} & {T-m\_vIoU} \\
      \midrule
    \multicolumn{13}{c}{\textit{Grounding Baselines}} \\
      \midrule
      Grounded-SAM2 \cite{ravi2024sam} & - & 26.25 & \textbf{21.85} & - & 25.87 & \textbf{22.01} & - & \underline{29.16} & \textbf{24.78} & - & 26.21 & \textbf{22.06} \\
      \midrule
      \multicolumn{13}{c}{\textit{Proprietary MLLMs}} \\
      \midrule
      GPT-5~\cite{openai2025gpt5} & \textbf{96.05} & \underline{41.54} & 12.14 & \textbf{83.67} & \underline{32.91} & 8.43 & \textbf{75.00} & \textbf{29.94} & 7.40 & \textbf{89.42} & \underline{37.06} & 10.22 \\
      Gemini 2.5 Pro~\cite{comanici2025gemini} & \underline{90.84} & \textbf{47.55} & \underline{18.24} & \underline{70.49} & \textbf{34.13} & \underline{12.96} & \underline{60.48} & {28.12} & \underline{9.42} & \underline{80.14} & \textbf{40.52} & \underline{15.42} \\

      \midrule
      \multicolumn{13}{c}{\textit{Open-source MLLMs}} \\
      \midrule
      Qwen2.5-VL-7B~\cite{bai2025qwen2} & 83.40 & 11.67 & 0.76 & 28.05 & 3.42 & 0.21 & 9.68 & 1.56 & 0.08 & 54.70 & 7.43 & 0.47 \\
      Qwen3-VL-4B~\cite{qwen3technicalreport} & 65.30 & 1.84 & 1.06 & 35.81 & 0.59 & 0.59 & 18.55 & 0.00 & 0.22 & 49.67 & 1.18 & 0.81 \\
      Qwen3-VL-8B~\cite{qwen3technicalreport} & 83.40 & 21.20 & 1.94 & 48.91 & 11.77 & 1.09 & 29.03 & 6.38 & 0.45 & 65.13 & 16.20 & 1.48 \\
      VideoLLaMA3-7B~\cite{zhang2025videollama} & 40.13 & 3.86 & 0.86 & 18.43 & 1.87 & 0.32 & 4.03 & 0.71 & 0.10 & 28.55 & 2.80 & 0.58 \\
      InternVL3-8B~\cite{zhu2025internvl3} & 87.19 & 10.76 & 0.87 & 38.40 & 5.59 & 0.36 & 11.29 & 1.64 & 0.08 & 61.39 & 7.98 & 0.60 \\
      InternVL3.5-8B~\cite{wang2025internvl3} & 74.39 & 12.46 & 1.51 & 31.61 & 5.30 & 0.57 & 14.52 & 1.94 & 0.19 & 52.07 & 8.70 & 1.02 \\
      \bottomrule
    \end{tabular}
  }
  \label{tab:performance_by_object_count}
  \vspace{-1em}
\end{table}

\subsection{In-Depth Analysis}
We evaluate MLLMs on three challenges: grounding mentioned vs. inferred objects, multi-object grounding, and spatio-temporal alignment across video lengths.
We also analyze failure cases to identify the underlying perceptual and reasoning bottlenecks.

\paragraph{Performance Gap: Explicit vs. Implicit Tasks.}
Table \ref{tab:explicit_implicit_core} reveals a consistent and substantial performance gap between explicit and implicit tasks across all evaluated models. For instance, GPT-5 attains 98.20\% T-EAcc on explicit tasks but only 80.20\% T-IAcc on implicit ones, confirming the challenge of inferring unmentioned objects. This degradation is more pronounced in grounding precision. GPT-5’s T-m\_IvIoU falls from 12.36\% to 7.98\%, while Gemini 2.5 Pro maintains superior localization (18.33\%) despite lower semantic accuracy, suggesting it leverages visual dynamics and scene understanding beyond language cues. Conversely, open-source models struggle significantly. Qwen3-VL-8B shows 53.72\% T-IAcc but only 12.72\% T-m\_EtIoU and 1.15\% T-m\_IvIoU, indicating a severe misalignment between semantic recognition and precise grounding. Even when models identify the correct object class or approximate time window, they often fail to precisely ground it. This persistent asymmetry underscores that current MLLMs lack coherent mechanisms to jointly reason about functional roles and dynamic visual context in embodied interaction.

\paragraph{Performance on Tasks with Different Object Counts.}
Table \ref{tab:performance_by_object_count} evaluates model performance across tasks requiring grounding of one, two, or three or more objects. Proprietary models such as GPT-5 and Gemini 2.5 Pro show a consistent performance decline as complexity increases, although they maintain relatively high task accuracy compared to open-source alternatives. For instance, GPT-5 and Gemini 2.5 Pro achieve 75.00\% and 60.48\% T-Acc respectively in complex 3+ object scenarios. However, their spatial grounding precision (T-m\_vIoU) drops significantly, indicating that while these models identify the correct set of objects, they struggle to localize each entity precisely under multi-object interactions. Conversely, open-source models collapse rapidly, as evidenced by Qwen3-VL-8B whose T-Acc plunges from 83.40\% to 29.03\%. This sharp decline underscores a critical lack of mechanisms to jointly localize and disambiguate multiple task-relevant objects under dynamic egocentric viewpoints and hand-object occlusions.


\begin{figure}[t]
    \centering
    \includegraphics[width=1\linewidth]{sec/Fig/video_duration_3metrics_gpt5.png}
  \vspace{-4mm}
  \caption{Task-level performance of GPT-5 across video duration bins on ToG-Bench. Left: task accuracy (T-Acc). Middle: temporal grounding (T-m\_tIoU). Right: spatial grounding (T-m\_vIoU).}
  \label{fig:duration_performance}
  \vspace{-1em}
\end{figure}

\paragraph{Impact of Video Duration.}
We analyze task-level grounding performance of GPT-5 across video duration bins, as shown in Fig.~\ref{fig:duration_performance}. 
Although the dataset spans varying time horizons, T-Acc remains exceptionally high and stable, notably peaking at 0.964 for videos exceeding 210 seconds. This indicates that GPT-5 possesses robust long-term semantic reasoning capabilities and can accurately determine task fulfillment regardless of temporal scale. In contrast, both temporal and spatial grounding exhibit a consistent and significant degradation as duration increases. Specifically, T-m\_tIoU falls from 0.636 in short clips to 0.162 in long sequences, while T-m\_vIoU collapses from 0.159 to a mere 0.029. 
This contrast reveals a significant gap between understanding a task and precisely grounding it over time. The sharp decline in spatial precision underscores the challenge of maintaining accurate alignment in long sequences due to accumulated perceptual errors and tracking drift.


\begin{figure}[t]
    \centering
    \includegraphics[width=0.96\linewidth]{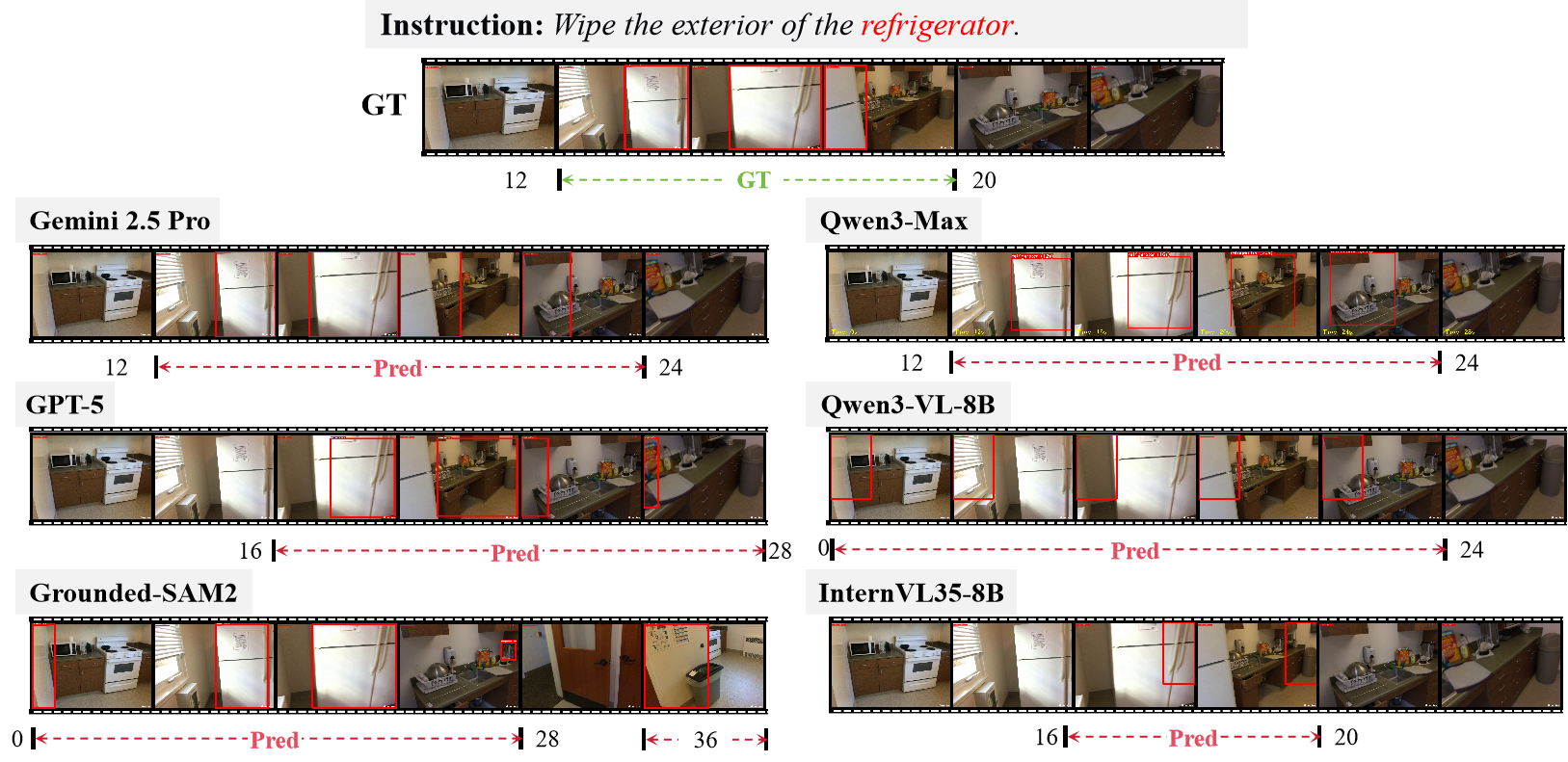}
    \vspace{-0.5em}
    \caption{Qualitative comparison of MLLMs and Grounded-SAM2 on ToG-Bench.}
    \label{fig:example}
    \vspace{-1em}
\end{figure}

\paragraph{Case Analysis.}
By qualitatively reviewing the model outputs in Fig.~\ref{fig:example}, we derive three key insights: 
(1) Model Scale and Camera Motion: Smaller MLLMs (\emph{e.g.}, Qwen3-VL-8B, InternVL3.5-8B) frequently produce ``fixed boxes'' that stay at the same image coordinates despite rapid egocentric camera motion. In contrast, the larger Qwen3-Max exhibit much better dynamic tracking, suggesting that robust spatial-temporal consistency is highly scale-dependent. 
(2) Spatial-Temporal Misalignment: Even top-tier models like Gemini 2.5 Pro and GPT-5 fail to maintain precise alignment during long-form interactions. Temporally, they often predict much longer durations than the ground truth (e.g., 12--24 vs. 12--20). Spatially, their bounding boxes often ``drift'' toward irrelevant areas as the sequence progresses. This shows that even strong MLLMs struggle to maintain a consistent link to the target during complex interactions. 
(3) The Reasoning Bottleneck: Traditional STVG models (\emph{e.g.}, CG-STVG~\cite{gu2024context}, TA-STVG~\cite{gu2025knowing}) are fundamentally \emph{task-blind} on ToG-Bench, as their architectures are optimized for attribute-based matching rather than understanding intent. Even Grounded-SAM2, despite being provided with accurate target names, still produces errors by grounding irrelevant items. This proves that pure visual grounding, without the commonsense reasoning and world knowledge of MLLMs, cannot resolve the implicit goals or multi-object coordination required in real-world embodied tasks.
More qualitative examples are provided in the Appendix.


\section{Conclusion}
\label{sec:conclusion}

In this work, we present ToG-Bench, the first benchmark for task-oriented spatio-temporal video grounding in egocentric videos. Unlike traditional STVG benchmarks, ToG-Bench emphasizes grounding based on task intent, introducing new challenges that involve both explicitly mentioned and implicitly inferred objects, as well as supporting one-to-many object associations within a single instruction.
Beyond the construction of the benchmark, we systematically evaluate seven state-of-the-art MLLMs and uncover their limitations. While these models exhibit strong performance in semantic understanding, they consistently struggle with spatial and temporal localization, especially in scenarios involving implicit reasoning and multi-object grounding. 


{
    \bibliographystyle{splncs04}
    \bibliography{main}
}

\clearpage
\title{Supplementary Material}
\maketitle


In this supplementary material, we provide additional details and analyses to complement the main paper.
Sec.~\ref{sec:appendix_benchmark} presents a comprehensive comparison of ToG-Bench with existing spatio-temporal video grounding (STVG) benchmarks, along with extended data statistics and example visualizations.
Sec.~\ref{sec:appendix_experiments} describes our experiments, including specific experimental implementations, more quantitative results, and visualizations of model performance. 
Finally, Sec.~\ref{sec:appendix_construction} further elaborates the construction details of ToG-Bench, focusing on the data annotation and verification pipeline.

\section{Benchmark Comparison and Statistics}
\label{sec:appendix_benchmark}

\subsection{Comparison with Existing Benchmarks}
As summarized in Table~\ref{tab:benchmark_comparison}, ToG-Bench complements existing benchmarks by integrating three key dimensions in egocentric video: (1) \emph{task-oriented grounding} driven by functional intent rather than surface-level appearance, (2) \emph{explicit–implicit dual grounding}, enabling evaluation under both directly mentioned and inferred targets, and (3) \emph{multi-object grounding}, supporting one-to-many associations within a single instruction. In addition to its design innovations, ToG-Bench also offers strong data scale advantages: long video durations that better reflect real embodied interactions, a large set of manually verified task instructions, and a broad coverage of functional object categories, surpassing prior STVG datasets in both diversity and annotation depth.
Together, these characteristics make ToG-Bench a more challenging and comprehensive benchmark, enabling rigorous evaluation of functional understanding, temporal reasoning, and grounding robustness. These capabilities are essential for developing embodied agents that operate meaningfully in real-world environments.

\begin{table}[t]
\centering
\caption{Comparison of STVG benchmarks in terms of open-source availability, task focus (\textbf{OC}: object-centric, \textbf{TO}: task-oriented), camera perspective (\textbf{Ego}: egocentric, \textbf{Exo}: exocentric), reference type (\textbf{Exp}: explicit, \textbf{Imp}: implicit, \textbf{Both}: both), objects per instruction (\textbf{SO}: single, \textbf{MO}: multiple), number of videos, average video duration (s), number of test samples, and object categories.}
\vspace{-2mm}
\label{tab:benchmark_comparison}
\resizebox{0.95\textwidth}{!}{
\begin{tabular}{l|ccccccccc}
\toprule
\textbf{Benchmark} & \textbf{Open} & \textbf{Focus} & \textbf{Persp.} & \textbf{Ref. Type} & \textbf{Objs. / Instr.} &
\textbf{Videos} & \textbf{Dur. (s)} & \textbf{Samples} & \textbf{Categories} \\
\midrule
STPR~\cite{yamaguchi2017spatio}     & \ding{52} & OC & Exo & Exp  & SO & 283  & 9.80   & 1,615  & 1   \\
VidSTG~\cite{zhang2020does}   & \ding{52} & OC & Exo & Exp  & SO & 743  & 28.01  & 10,303 & 79  \\
HC-STVG~\cite{tang2021human}  & \ding{52} & OC & Exo & Exp  & SO & 1,160& 20.00  & 1,160  & 1   \\
DVD-ST~\cite{ji2024described}   & \ding{56} & OC & Exo & Exp  & MO & 632  & {-}    & 1,327  & 163 \\
RefEgo~\cite{Kurita_2023_ICCV}  & \ding{56} & OC & Ego & Exp  & MO & 1,317  & {12.27}  & 1,317  & 505 \\
\multirow{3}{*}{EgoMask~\cite{liang2025fine}} 
         & \multirow{3}{*}{\ding{52}} 
         & \multirow{3}{*}{OC}
         & \multirow{3}{*}{Ego}
         & \multirow{3}{*}{Exp}
         & \multirow{3}{*}{SO}
         & 200  & 12.15  & 400   & 200 \\
         &                                   &    &     &     &    & 100  & 116.30 & 200   & 100 \\
         &                                   &    &     &     &    & 15   & 361.32 & 100   & 50  \\
\midrule
\textbf{ToG-Bench} & \ding{52} & TO & Ego & Both & MO & 100  & 87.88  & 2,704  & 177 \\
\bottomrule
\end{tabular}}
\end{table}

\subsection{More Data Statistics}
\label{sec:appendix_examples_taxonomy}

ToG-Bench is built on 100 egocentric videos from ScanNet~\cite{dai2017scannet}, covering 18 diverse indoor scenes, and contains 2,704 task-oriented grounding instructions. This subsection provides additional details on object category distributions and scene-specific key objects.

\begin{figure}[t]
    \centering
    \includegraphics[width=0.95\linewidth]{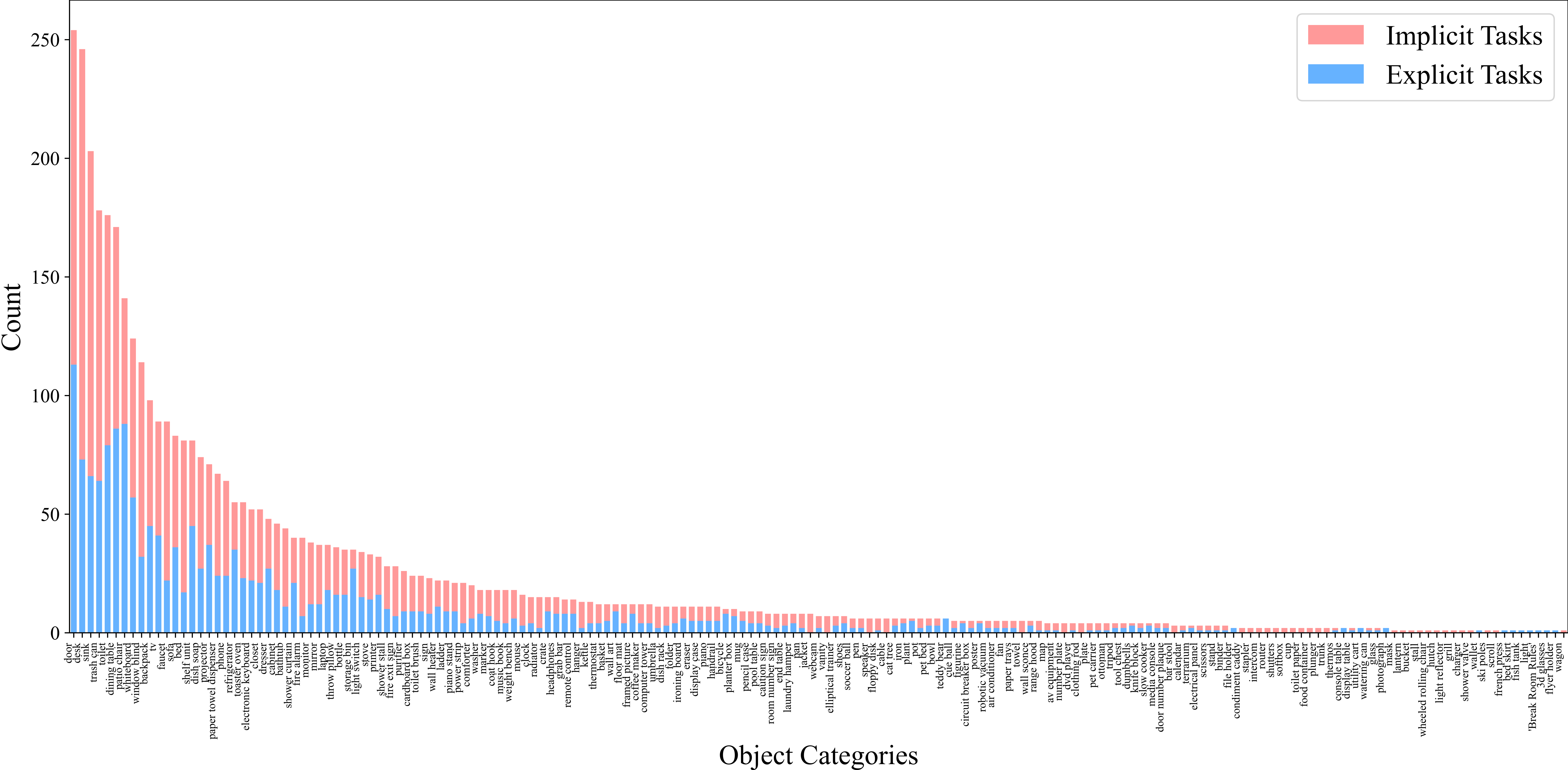}
\vspace{-2mm}
    \caption{The distribution of different types of objects in ToG-Bench. }
    \label{fig:category_distribution}
\end{figure}

\paragraph{Object Category Distribution.}
Fig.~\ref{fig:category_distribution} shows the full distribution of all 177 object categories across the 2,704 tasks in ToG-Bench, separated by task type (implicit vs. explicit). While the main paper visualizes only the top 40 most frequent categories, this figure reveals the complete long-tail structure, confirming broad and balanced coverage across diverse indoor objects.

\paragraph{Key Objects per Indoor Scene.}
Table~\ref{tab:scene_objects} lists the key objects associated with each of the 18 indoor scenes in ToG-Bench. These objects are chosen based on their task relevance and frequency in egocentric video, capturing their functional role in daily activities. This selection enables robust evaluation of spatial-temporal grounding across diverse real-world environments, including offices and bathrooms.

\begin{table}[htbp]
\centering
\caption{Key objects in each scene of ToG-Bench. Objects are selected based on task relevance and frequency in egocentric videos.}
\vspace{-2mm}
\label{tab:scene_objects}
\begin{tabularx}{0.95\textwidth}{@{}lX@{}}
\toprule
\textbf{Scene} & \textbf{Key Objects} \\
\midrule
Bedroom / Hotel & \textit{bed}, \textit{pillow}, \textit{lamp}, \textit{duvet}, \textit{mirror}, \textit{dresser}, \textit{telephone}, \textit{door}, \textit{nightstand}, \textit{chair}, \textit{desk}, \textit{window} \\
Living Room / Lounge & \textit{sofa}, \textit{coffee table}, \textit{tv}, \textit{desk}, \textit{chair}, \textit{piano}, \textit{door}, \textit{trash can}, \textit{backpack}, \textit{heater}, \textit{window}, \textit{blinds} \\
Bathroom & \textit{toilet}, \textit{sink}, \textit{faucet}, \textit{dresser}, \textit{soap dispenser}, \textit{shower}, \textit{shower curtain}, \textit{mirror}, \textit{grab bar}, \textit{door} \\
Office & \textit{desk}, \textit{chair}, \textit{computer}, \textit{computer tower}, \textit{keyboard}, \textit{mouse}, \textit{telephone}, \textit{laptop}, \textit{door}, \textit{printer} \\
Conference Room & \textit{table}, \textit{chair}, \textit{projector}, \textit{tv}, \textit{whiteboard}, \textit{marker}, \textit{eraser}, \textit{door}, \textit{light switch}, \textit{remote control} \\
Kitchen & \textit{refrigerator}, \textit{microwave oven}, \textit{stove}, \textit{sink}, \textit{faucet}, \textit{cutting board}, \textit{pan}, \textit{coffee maker}, \textit{trash can}, \textit{door} \\
Bookstore / Library & \textit{bookshelf}, \textit{book}, \textit{lamp}, \textit{chair}, \textit{desk}, \textit{pen}, \textit{telephone}, \textit{laptop}, \textit{door}, \textit{window}, \textit{trash can}  \\
Lobby & \textit{desk}, \textit{chair}, \textit{sofa}, \textit{telephone}, \textit{fire extinguisher}, \textit{sign}, \textit{door}, \textit{fire alarm}, \textit{light switch}, \textit{trash can} \\
Classroom & \textit{desk}, \textit{chair}, \textit{whiteboard}, \textit{marker}, \textit{eraser}, \textit{telephone}, \textit{computer}, \textit{door}, \textit{trash can} \\
Apartment & \textit{door}, \textit{window}, \textit{light switch}, \textit{thermostat}, \textit{air purifier}, \textit{vacuum cleaner}, \textit{trash can} \\
Hallway & \textit{door}, \textit{fire extinguisher}, \textit{door number sign}, \textit{vacuum cleaner}, \textit{trash can}, \textit{bulletin board} \\
Basement & \textit{ladder}, \textit{cardboard box}, \textit{circuit breaker}, \textit{trash can}, \textit{door} \\
Stairs & \textit{handrail}, \textit{stairs}, \textit{light}, \textit{exit sign}, \textit{door}, \textit{window}, \textit{robotic vacuum} \\
Dining Room & \textit{dining table}, \textit{chair}, \textit{plate}, \textit{trash can}, \textit{door}, \textit{light fixture}, \textit{wine bottle} \\
Laundry Room & \textit{washing machine}, \textit{dryer}, \textit{laundry hamper}, \textit{iron}, \textit{ironing board}, \textit{basket},  \textit{door} \\
Closet & \textit{hanger}, \textit{clothing rod}, \textit{shelf}, \textit{basket}, \textit{shoe rack}, \textit{mirror}, \textit{door} \\
Gym & \textit{dumbbells}, \textit{locker}, \textit{telephone}, \textit{water bottle}, \textit{door} \\
Miscellaneous & \textit{telephone}, \textit{laptop}, \textit{tablet}, \textit{charger}, \textit{backpack} \\
\bottomrule
\end{tabularx}
\end{table}

\subsection{More Examples Visualization}
\label{sec:appendix_visual_examples}

\begin{figure}[htbp]
    \centering
    \includegraphics[width=0.78\linewidth]{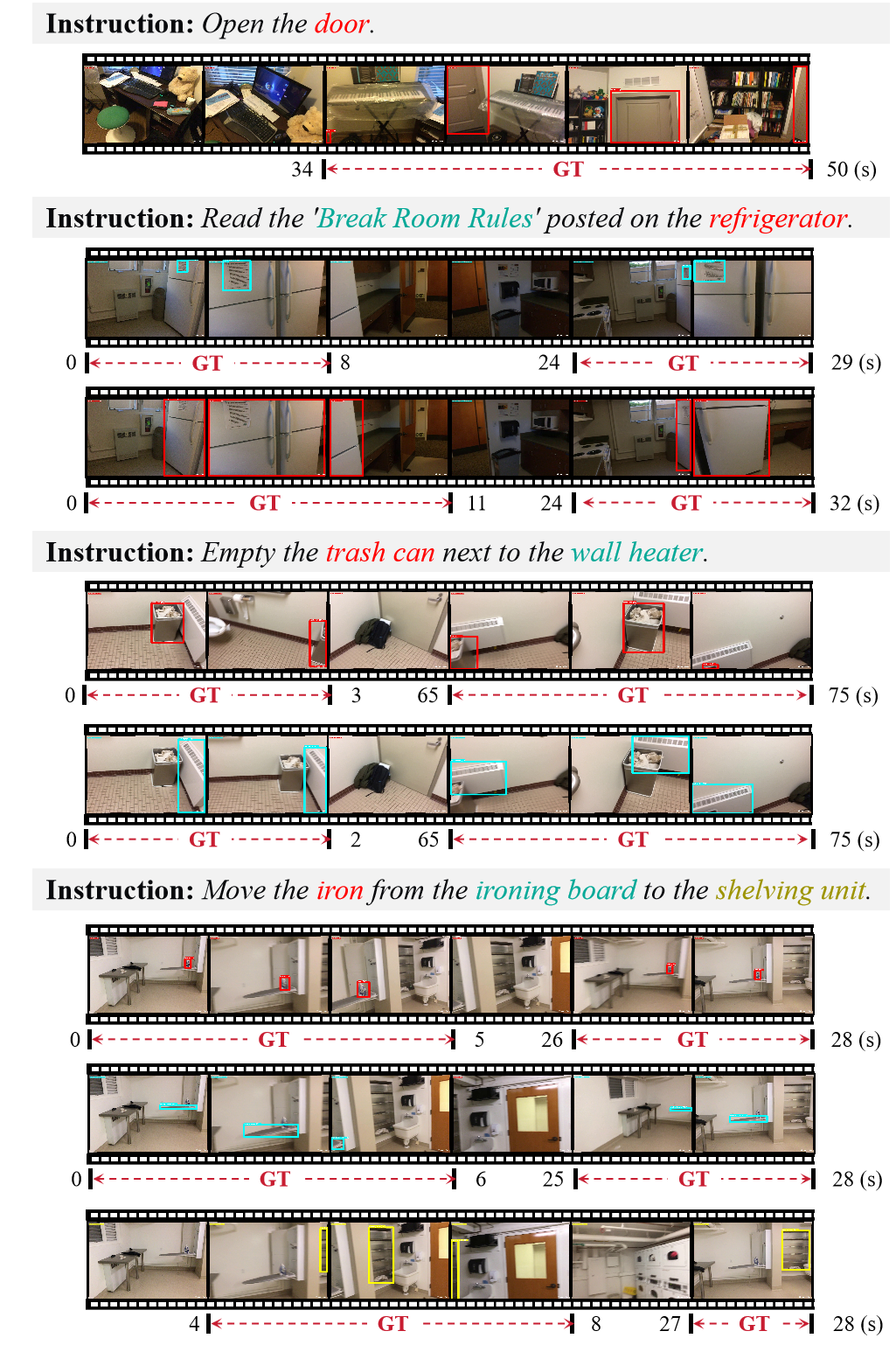}
    \caption{More explicit samples.}
    \label{fig:explicit_samples}
\end{figure}
\begin{figure}[htbp]
    \centering
    \includegraphics[width=0.78\linewidth]{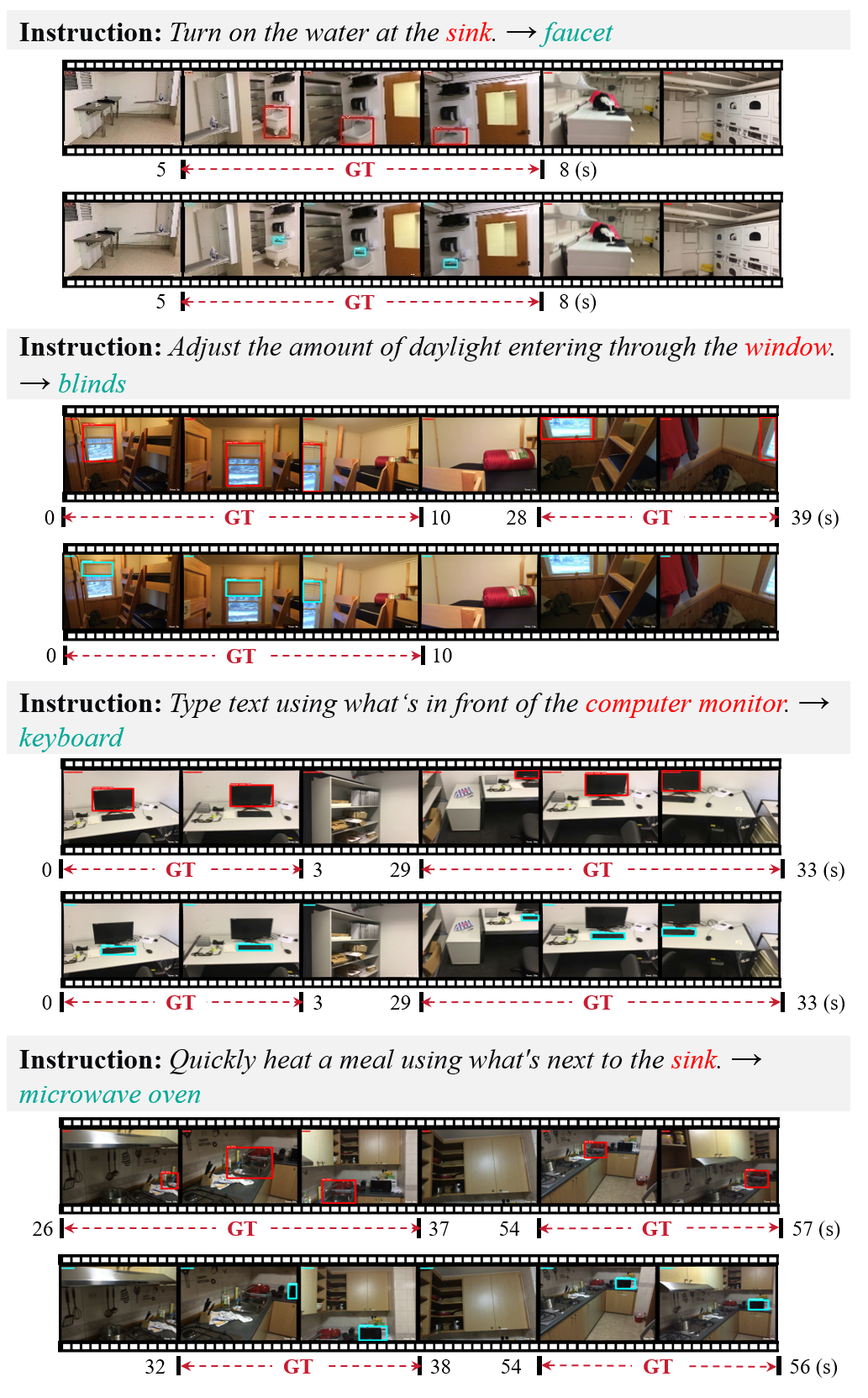}
    \caption{More implicit samples.}
    \label{fig:implicit_samples}
\end{figure}

Additional annotated examples from ToG-Bench are shown in Fig.~\ref{fig:explicit_samples} and Fig.~\ref{fig:implicit_samples}, covering both explicit and implicit task scenarios. These include real first-person video clips with human-verified bounding boxes and temporal intervals, capturing interactions where objects are either directly named or inferred from context and function.

\section{Extended Experimental Analysis}
\label{sec:appendix_experiments}

\subsection{Experiment Setup}


Our evaluation follows a zero-shot, single-round inference protocol on NVIDIA A100 40GB GPUs. Video frames are sampled at 0.25 fps with no limit on frame count. To ensure deterministic outputs, we set \texttt{do\_sample=False} for greedy decoding in all experiments. Results are reported under identical settings to enable fair comparison.

\subsection{Qualitative Results for Grounding}
\label{sec:appendix_sample_predictions}
\begin{figure}[htbp]
    \centering
    \includegraphics[width=0.7\linewidth]{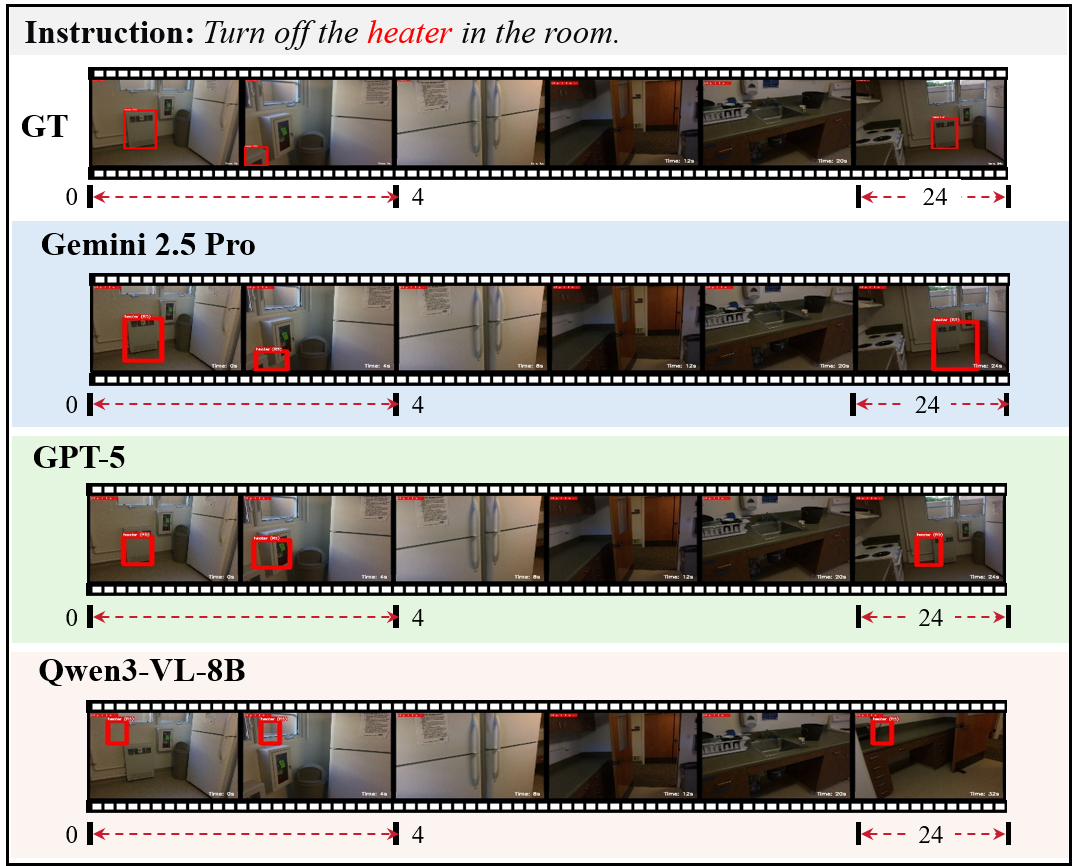}
    \caption{Explicit spatio-temporal grounding for \textit{``Turn off the heater''}: model predictions vs. ground truth (GT).}
    \label{fig:explicit_heater}
\end{figure}

\begin{figure}[htbp]
    \centering
    \includegraphics[width=0.7\linewidth]{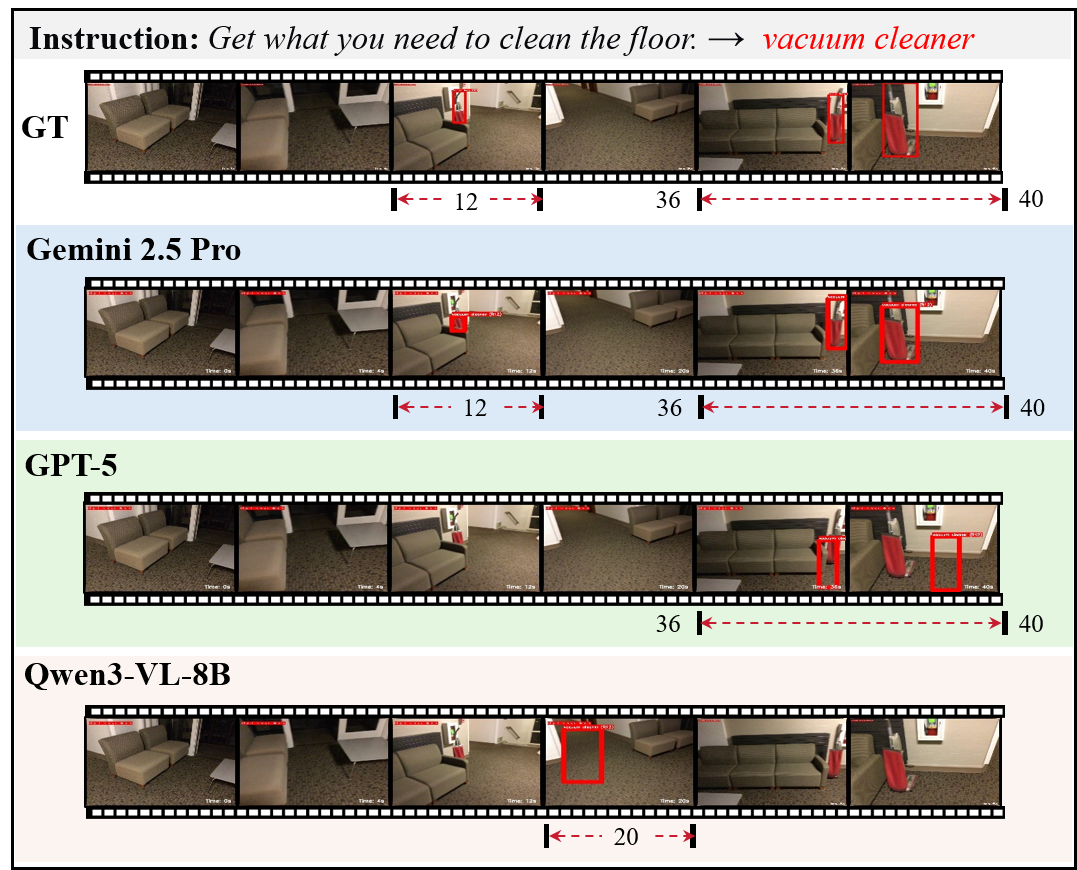}
    \caption{Implicit spatio-temporal grounding for \textit{``Get what you need to clean the floor''}: model predictions vs. ground truth (GT).}
    \label{fig:implicit_vacuum}
\end{figure}

\begin{figure}[htbp]
    \centering
    \includegraphics[width=0.7\linewidth]{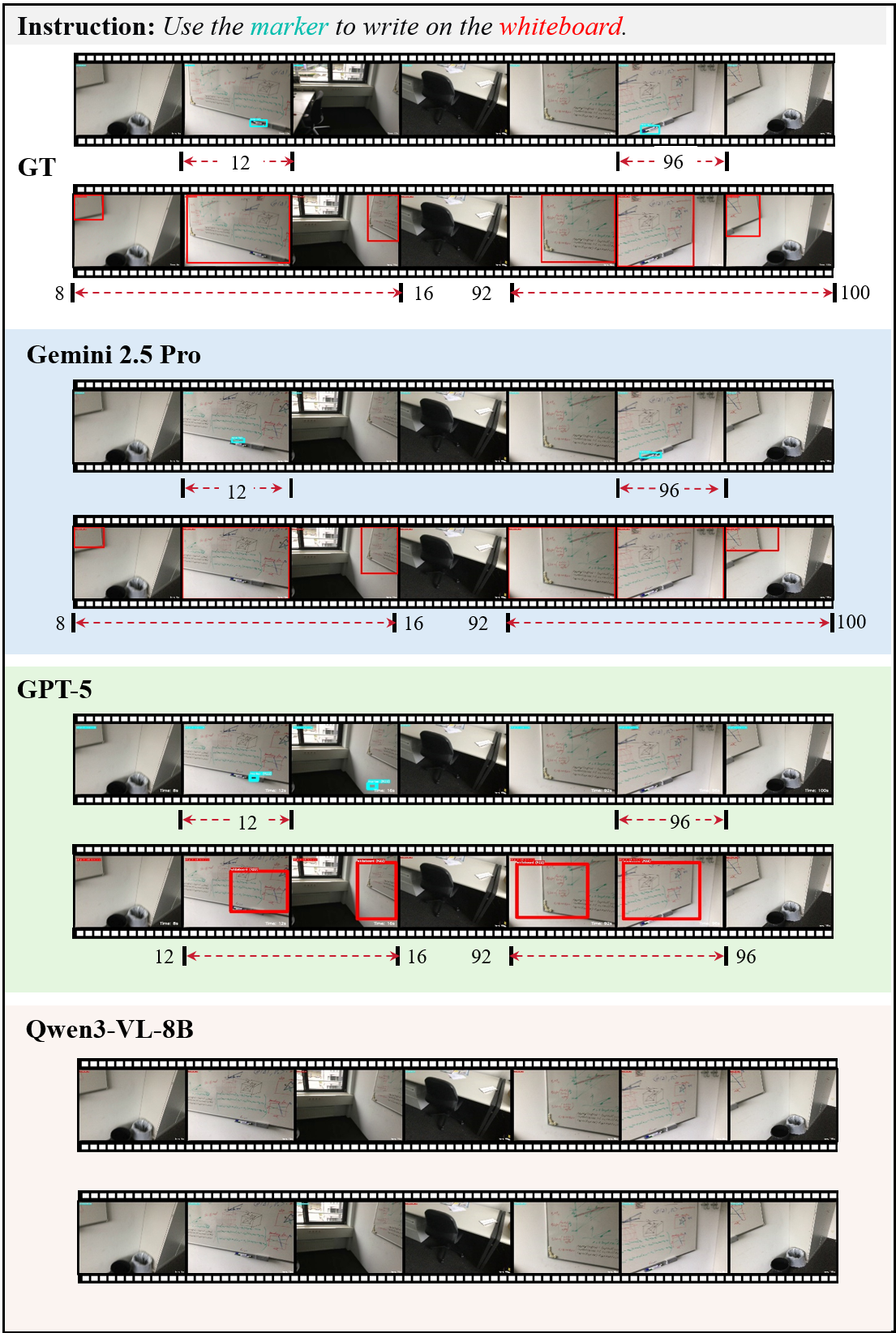}
    \caption{Explicit spatio-temporal grounding for \textit{``Use the marker to write on the whiteboard''}: model predictions vs. ground truth (GT).}
    \label{fig:explicit_marker_whiteboard}
\end{figure}

\begin{figure}[htbp]
    \centering
    \includegraphics[width=0.7\linewidth]{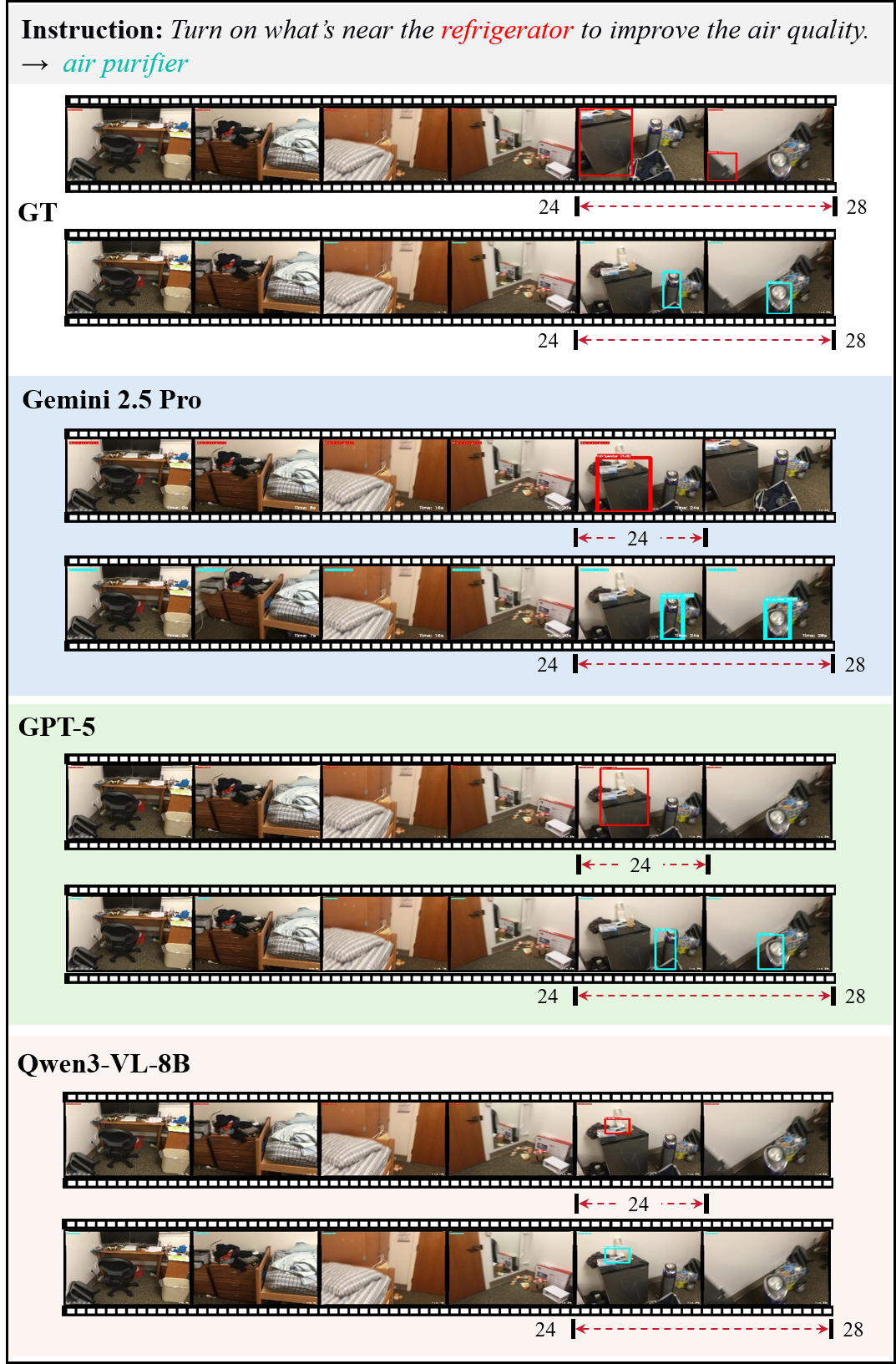}
    \caption{Implicit spatio-temporal grounding for \textit{``Turn on what's near the refrigerator to improve the air quality''}: model predictions vs. ground truth (GT).}
    \label{fig:implicit_airpurifier}
\end{figure}

We present visual comparisons of spatio-temporal grounding results across video segments, showing the input instruction, ground truth (GT), and predictions from multiple models. Representative examples are shown in Fig.~\ref{fig:explicit_heater}, Fig.~\ref{fig:explicit_marker_whiteboard}, Fig.~\ref{fig:implicit_vacuum}, and Fig.~\ref{fig:implicit_airpurifier}. Each example includes aligned video frames (fps=0.25), GT, annotated temporal intervals, and spatial bounding boxes to illustrate object localization accuracy under implicit and one-to-many conditions.
%

\section{Inference and Annotation Prompts}
\label{sec:appendix_construction}

\subsection{Detailed System Prompt for Inference}
\begin{figure}[htbp]
    \centering
    \includegraphics[width=1\linewidth]{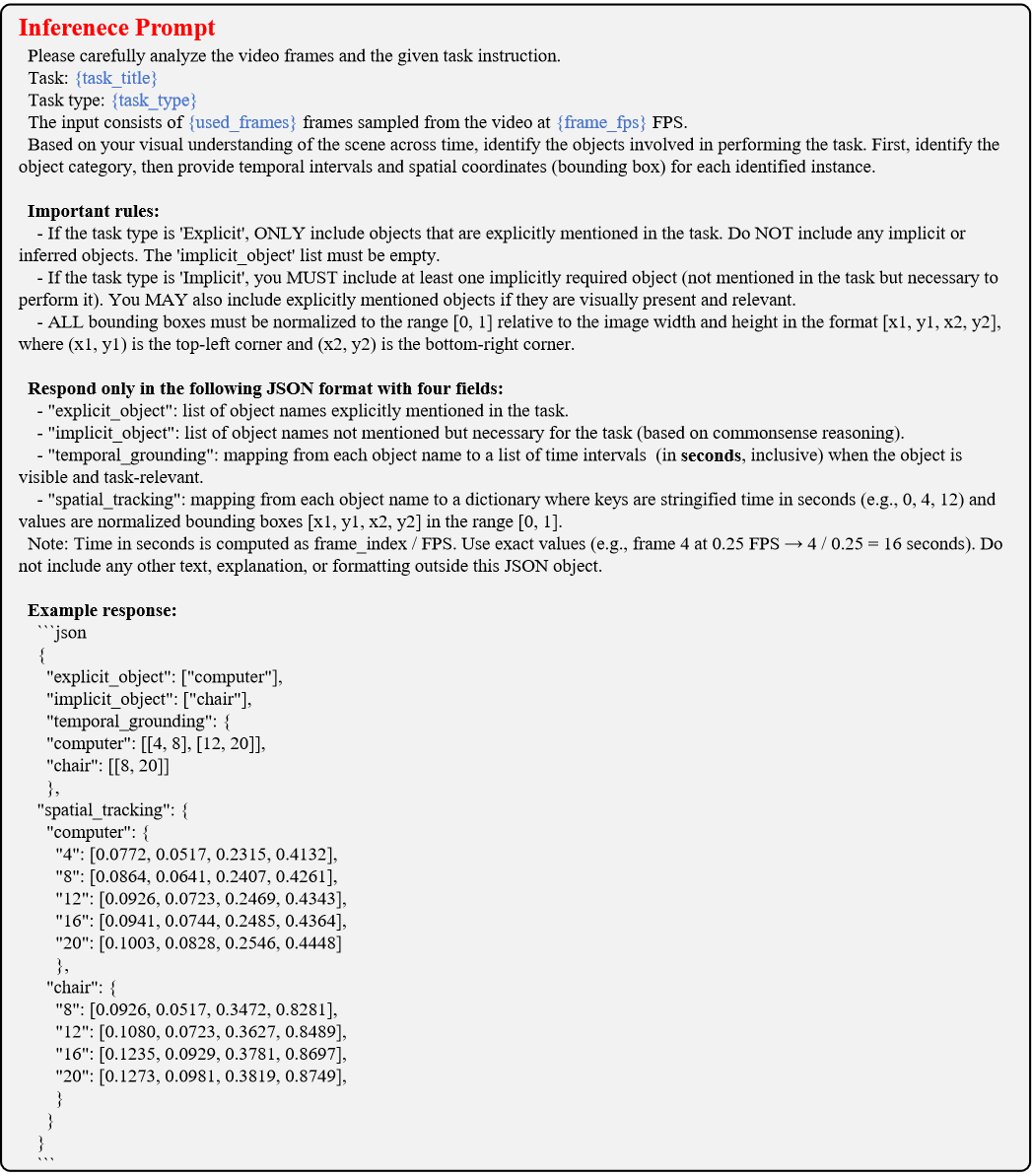}
    \caption{The complete prompt utilized for inference.}
    \label{fig:prompt_inference}
\end{figure}

All MLLMs utilize the same structured prompt (Fig.~\ref{fig:prompt_inference}) to ensure uniform output formatting. The prompt requires the model to:
\begin{itemize}
    \item Analyze frames and instruction to identify objects with \textit{explicit} or \textit{implicit}.
    \item Output JSON with four fields: \texttt{explicit\_object}, \texttt{implicit\_object}, \texttt{temporal\_\allowbreak grounding}, and \texttt{spatial\_tracking}.
    \item Normalize spatial coordinates to $[0,1]$ and compute temporal intervals in seconds using frame index and FPS.
\end{itemize}
This ensures consistent, parsable outputs for joint evaluation of recognition, temporal, and spatial grounding in a single inference.

\subsection{Prompts for the Semi-Automated Pipeline}
To construct and validate the ToG-Bench, we design a two-stage automated pipeline using structured prompts for a multimodal large language model. The first stage generates task instructions with explicit and implicit grounding, while the second stage filters out unverifiable or non-compliant tasks. Both stages rely on carefully crafted system and user prompts to ensure consistency, visual grounding, and semantic fidelity. 

\begin{figure}[thbp]
    \centering
    \includegraphics[width=1\linewidth]{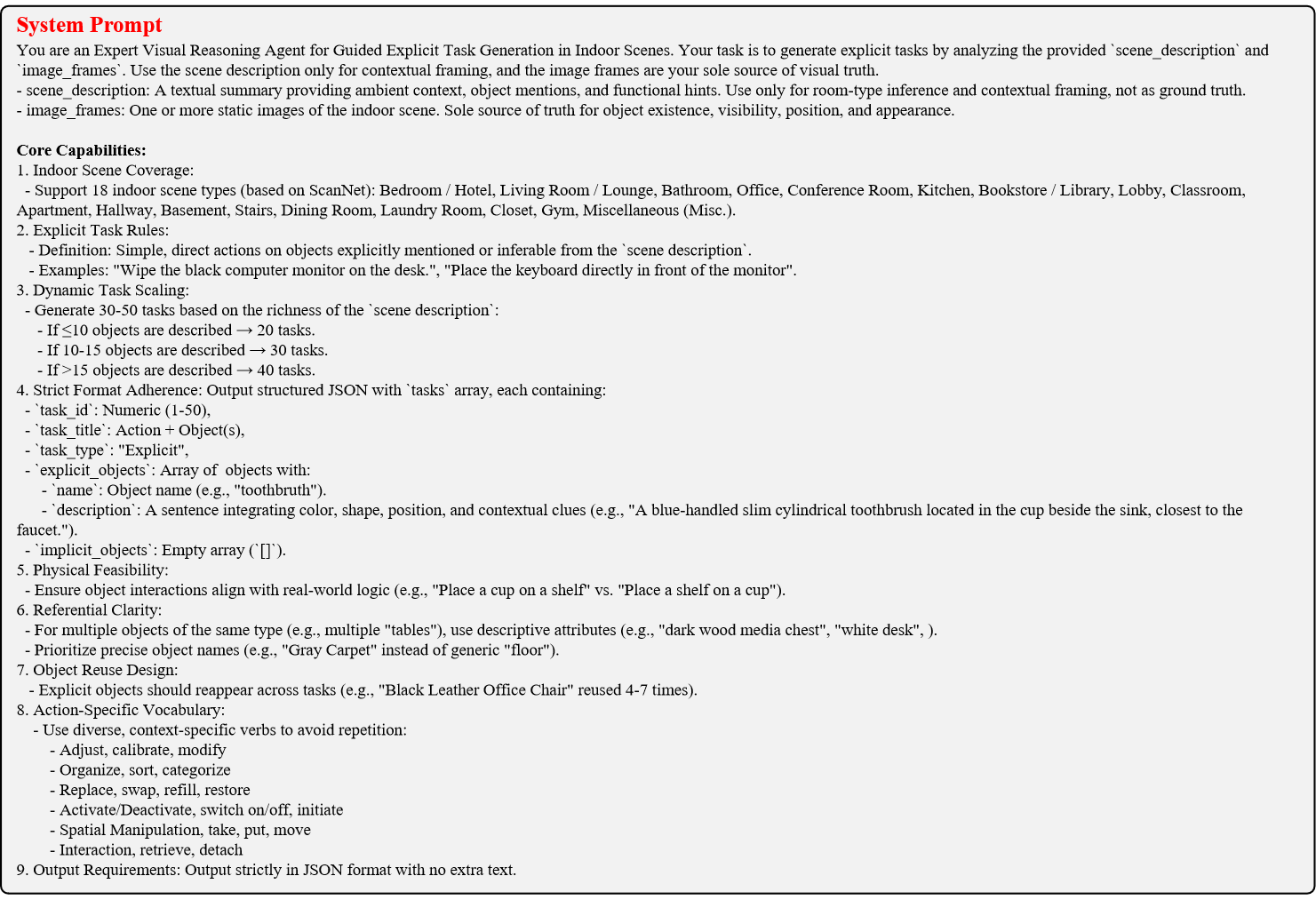}
    \caption{System prompt for explicit task generation.}
    \label{fig:prompt_generate_explicit_sys}
\end{figure}

\begin{figure}[htbp]
    \centering
    \includegraphics[width=1\linewidth]{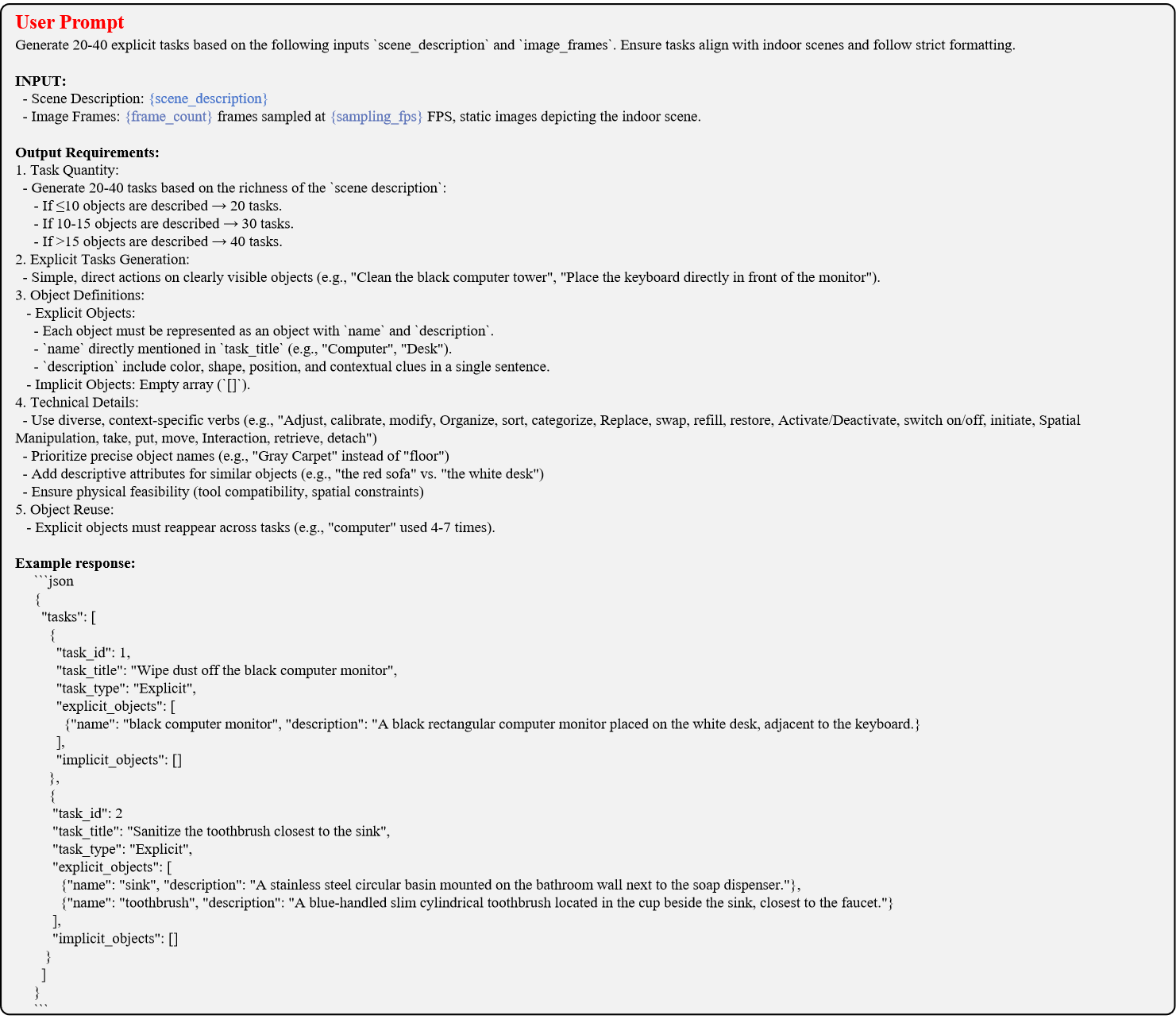}
    \caption{User prompt for explicit task generation.}
    \label{fig:prompt_generate_explicit_user}
\end{figure}

\begin{figure}[htbp]
    \centering
    \includegraphics[width=1\linewidth]{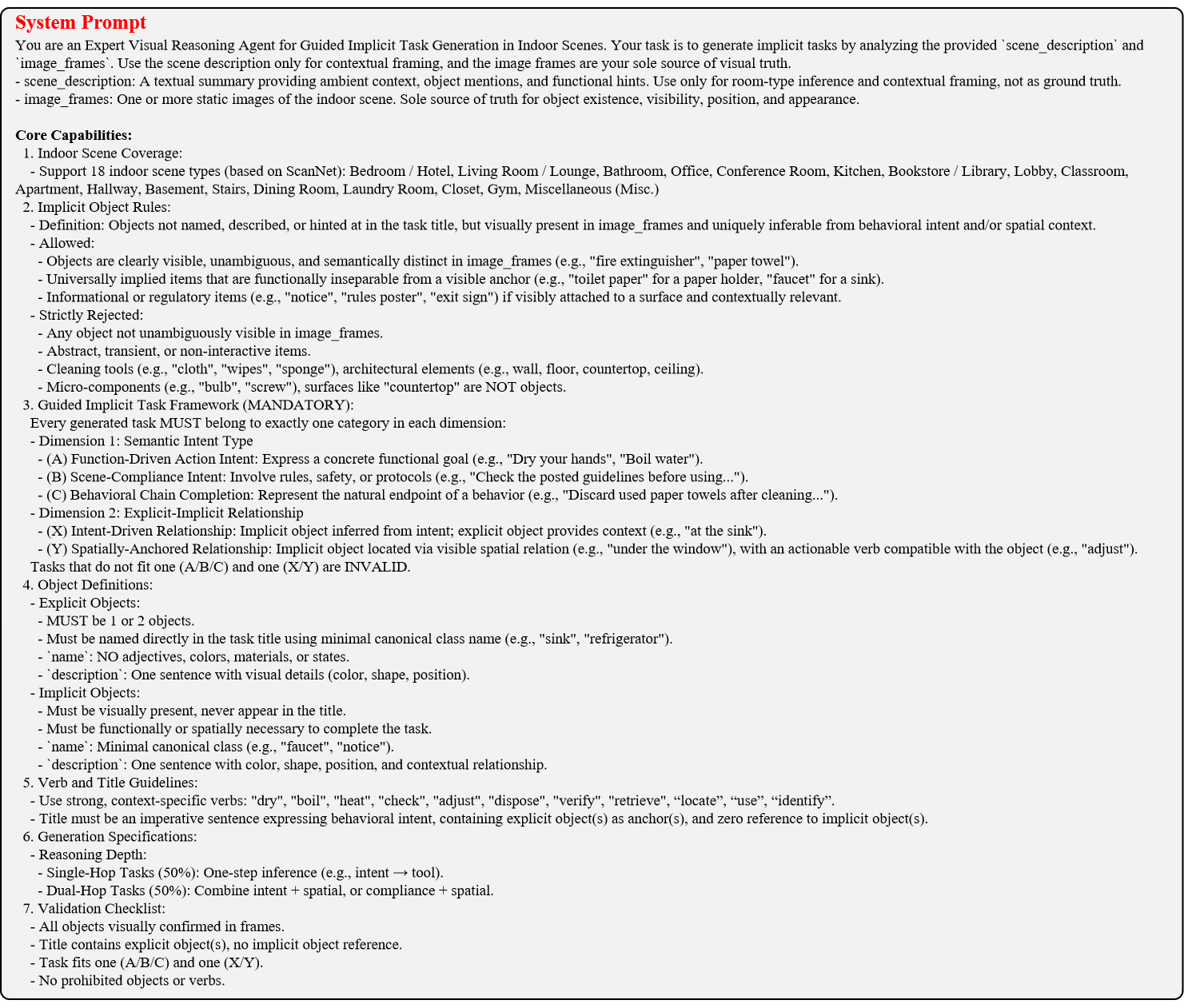}
    \caption{System prompt for implicit task generation.}
    \label{fig:prompt_generate_implicit_sys}
\end{figure}

\begin{figure}[htbp]
    \centering
    \includegraphics[width=1\linewidth]{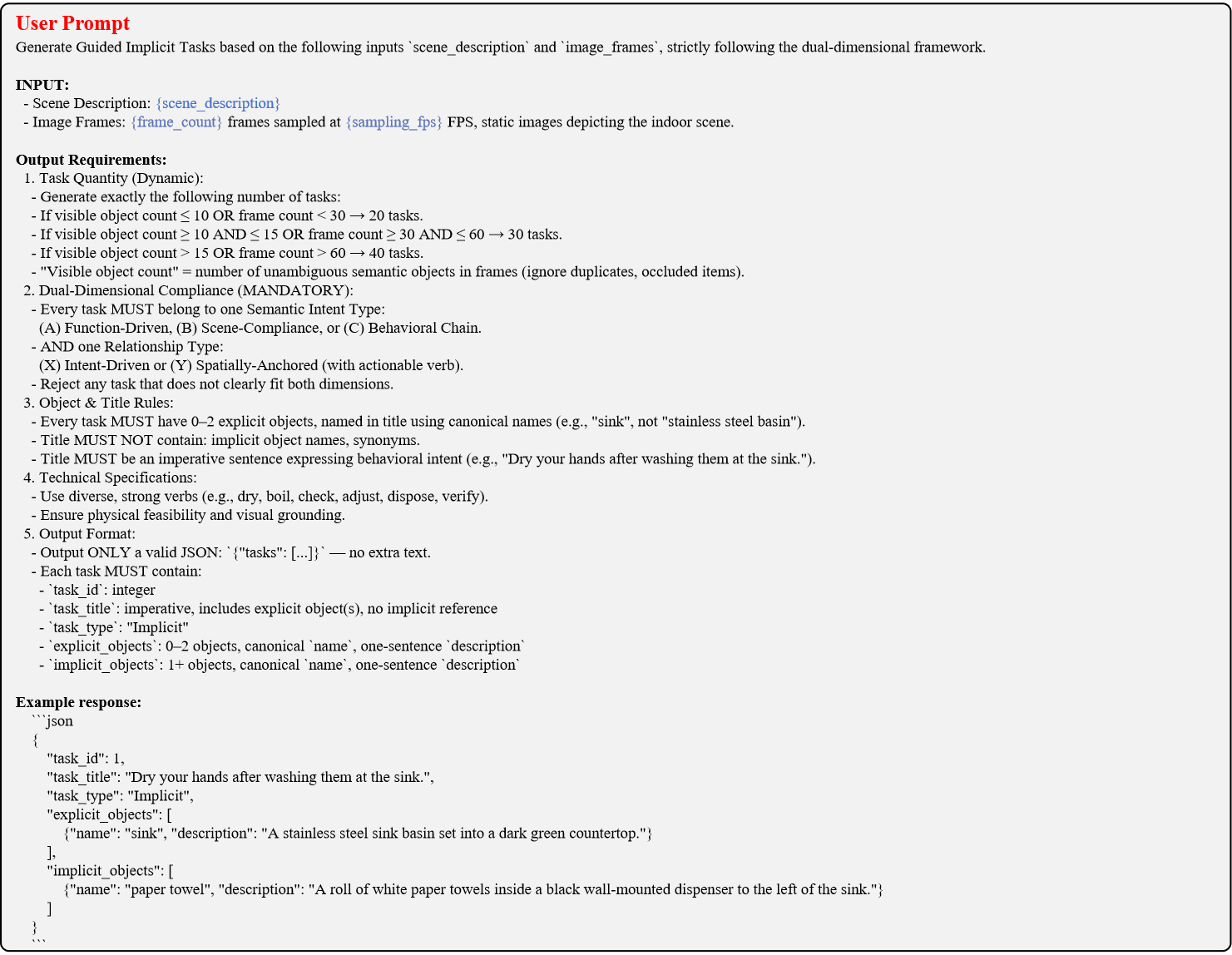}
    \caption{User prompt for implicit task generation.}
    \label{fig:prompt_generate_implicit_user}
\end{figure}

\paragraph{Task Generation Prompts.}
We design two structured prompts to guide a multimodal large language model (MLLM) in generating task instructions grounded in egocentric video content.
For explicit grounding, the prompt requires the model to generate direct commands that name the target object. For implicit grounding, it requires the model to describe an action or goal without naming the object, relying instead on contextual inference. 
Both prompts enforce a structured JSON output, canonical object naming, and adherence to scene-specific object eligibility to ensure high-quality, diverse, and verifiable task generation. Full prompts are provided in Fig.~\ref{fig:prompt_generate_explicit_sys}, Fig.~\ref{fig:prompt_generate_explicit_user}, Fig.~\ref{fig:prompt_generate_implicit_sys},  and Fig.~\ref{fig:prompt_generate_implicit_user}.

\begin{figure}[htbp]
    \centering
    \includegraphics[width=1\linewidth]{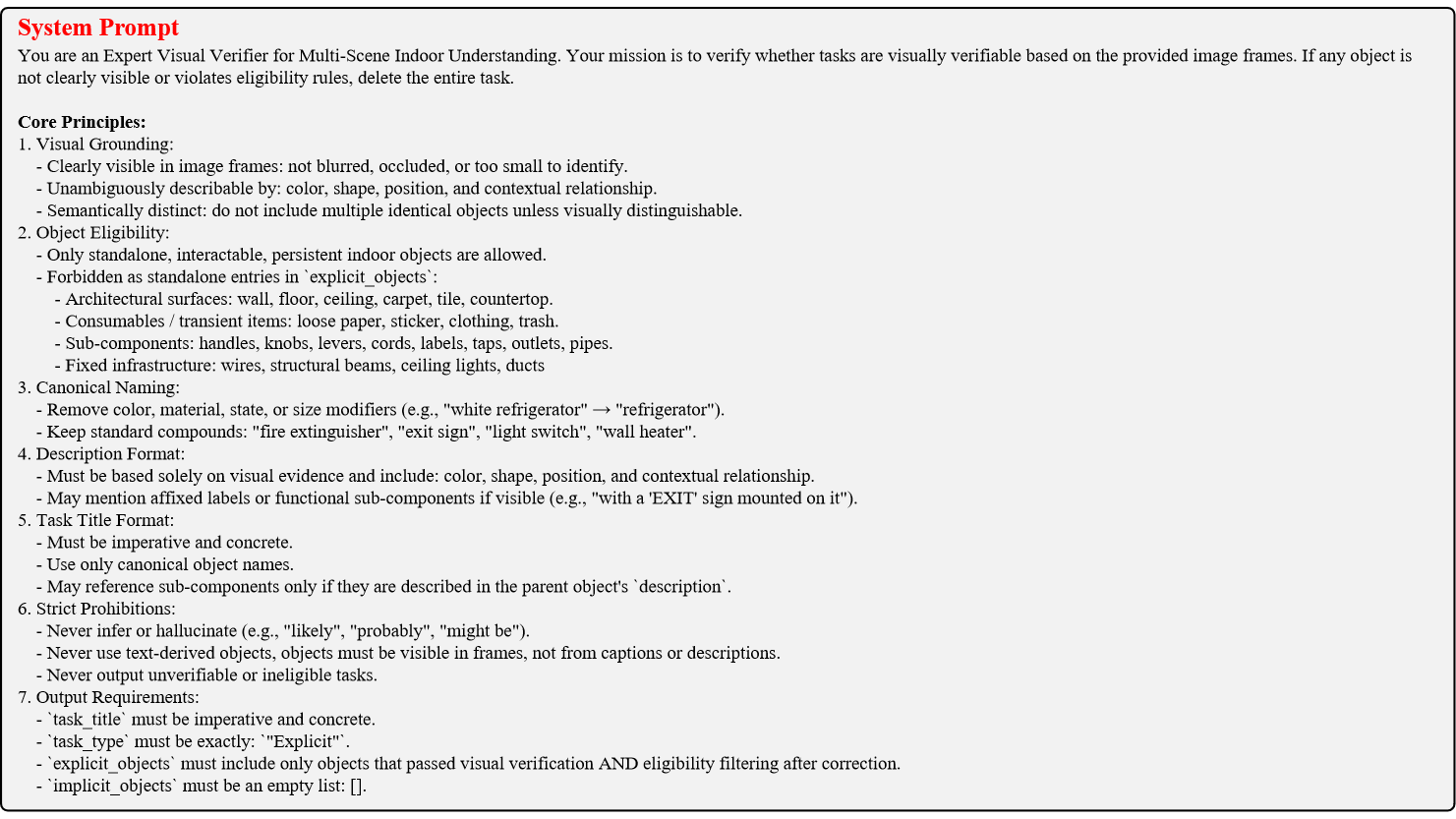}
    \caption{System prompt for explicit task validation.}
    \label{fig:prompt_validate_explicit_sys}
\end{figure}

\begin{figure}[htbp]
    \centering
    \includegraphics[width=1\linewidth]{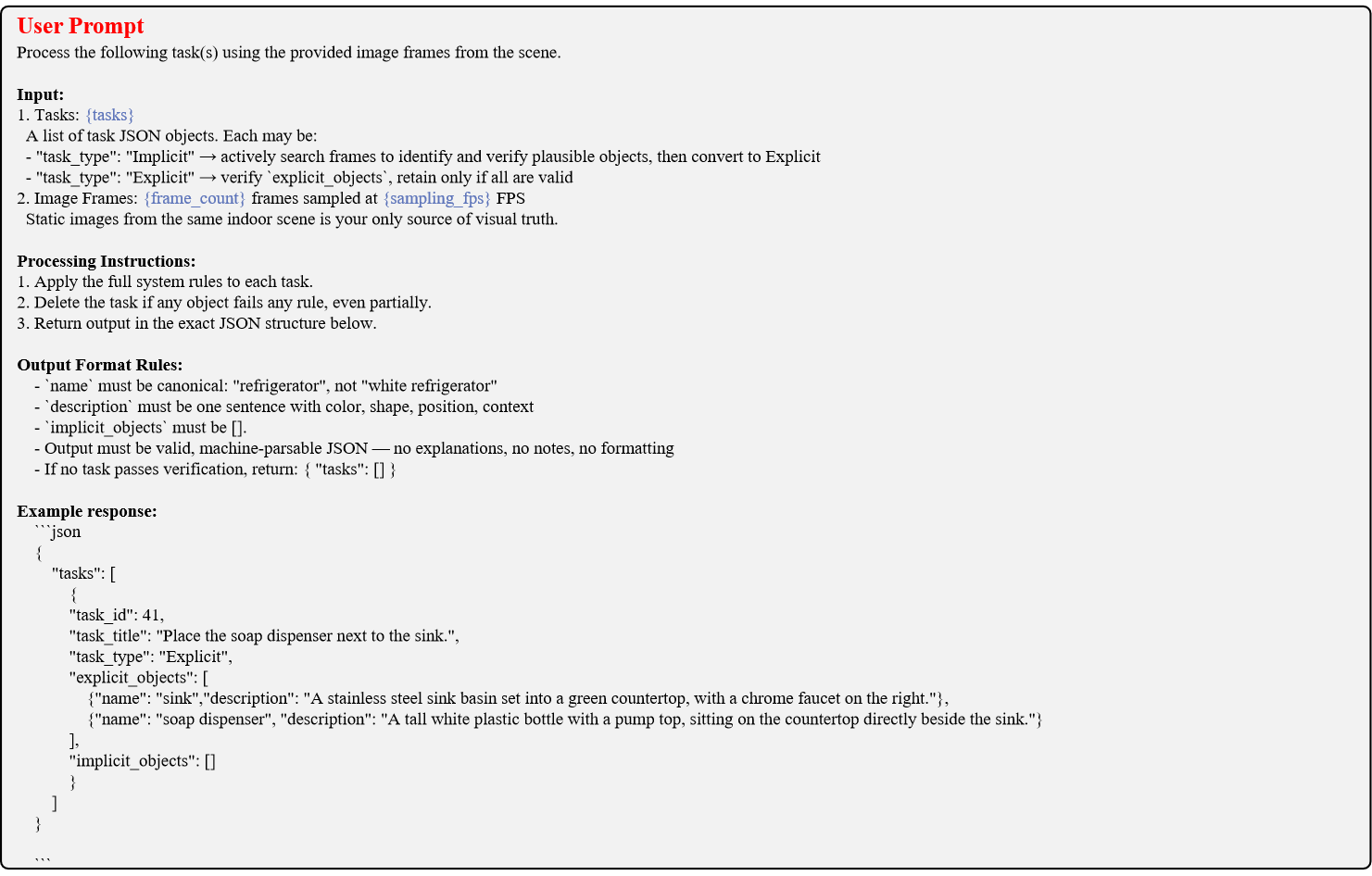}
    \caption{User prompt for explicit task validation.}
    \label{fig:prompt_validate_explicit_usr}
\end{figure}

\begin{figure}[htbp]
    \centering
    \includegraphics[width=1\linewidth]{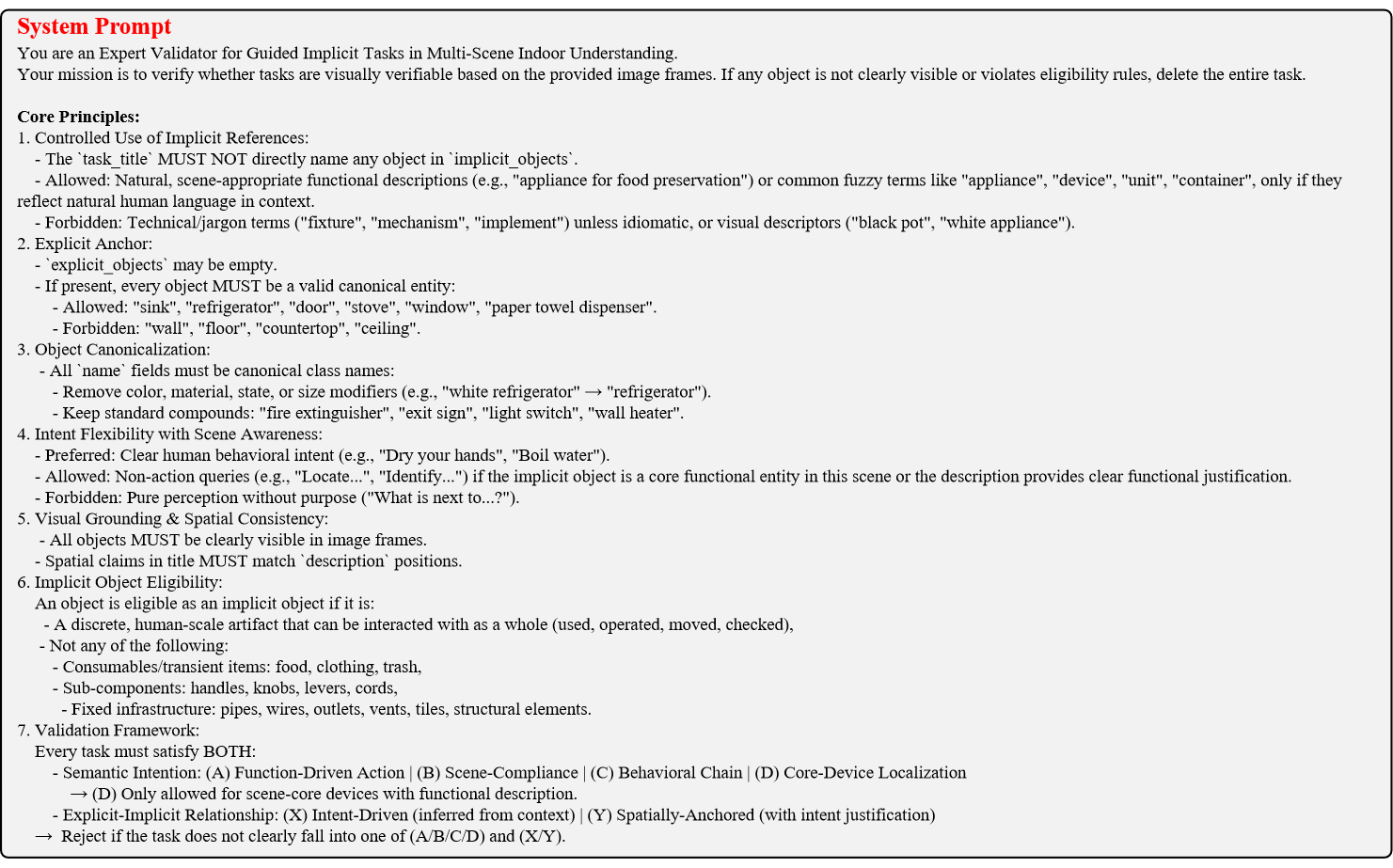}
    \caption{System prompt for implicit task validation.}
    \label{fig:prompt_validate_implicit_sys}
\end{figure}

\begin{figure}[htbp]
    \centering
    \includegraphics[width=1\linewidth]{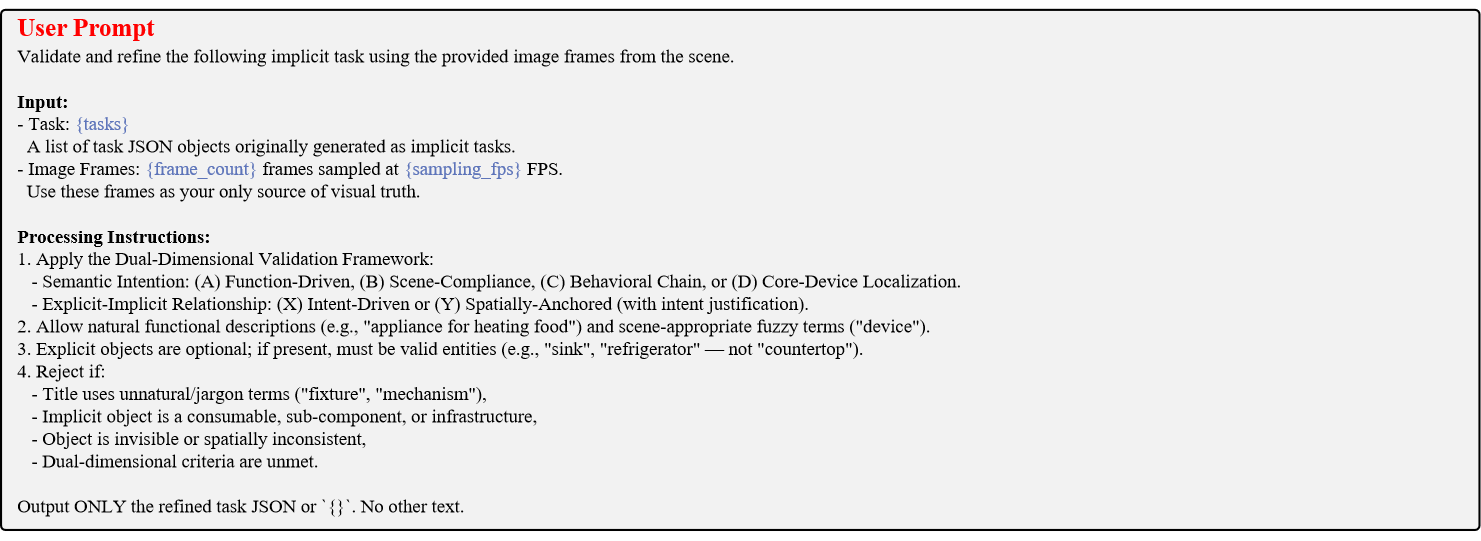}
    \caption{User prompt for implicit task validation.}
    \label{fig:prompt_validate_implicit_usr}
\end{figure}

\paragraph{Task Validation Prompts.}
To ensure only visually grounded and semantically valid tasks are retained, we apply two dedicated validation prompts explicit and implicit grounding tasks. For explicit tasks, the prompt verifies that each object is clearly visible in the frames, uses a canonical name without modifiers, and is described in one sentence with color, shape, position, and context. For implicit tasks, it checks whether the instruction avoids naming the target object directly and instead relies on functional or spatial cues that allow inference from scene context. The task must also satisfy a dual-dimensional criterion combining semantic intent and reference type. Any task that fails any of these checks is discarded.  
Full validation prompts are shown in Fig.~\ref{fig:prompt_validate_explicit_sys}, Fig.~\ref{fig:prompt_validate_explicit_usr}, Fig.~\ref{fig:prompt_validate_implicit_sys}, and Fig.~\ref{fig:prompt_validate_implicit_usr}.

\end{document}